\documentclass[mathpazo]{cicp}

\usepackage[utf8]{inputenc}

\usepackage{fullpage} 
\usepackage{multirow}
\usepackage{lineno}
\usepackage{enumerate}
\usepackage{xcolor}
\usepackage{tikz}
\usetikzlibrary{calc, positioning, shapes.arrows, arrows}
\usepackage{url}
\usepackage{stmaryrd}
\modulolinenumbers[5]

\usepackage{nomencl}
\usepackage{amsmath,mathtools}
\usepackage{gensymb}
\usepackage{amsfonts,bm}
\usepackage{multirow}
\usepackage{subfig}
\usepackage{longtable,tabularx}

\usepackage{graphicx}
\usepackage{booktabs}
\usepackage{makecell}
\usepackage[skip=0.6\baselineskip]{caption}
\usepackage{array}
\newcolumntype{P}[1]{>{\centering\arraybackslash}m{#1}}

\usepackage{xpatch}
\let\oldequation\equation
\let\oldendequation\endequation
\renewenvironment{equation}
  {\linenomathNonumbers\oldequation}
  {\oldendequation\endlinenomath}
  
\renewcommand\nomgroup[1]{%
  \item[\itshape
  \ifstrequal{#1}{A}{Symbols}{%
  \ifstrequal{#1}{B}{Roman Letters}{%
  \ifstrequal{#1}{C}{Greek Letters}{%
  \ifstrequal{#1}{D}{Abbreviations}{}}}}%
]}

\newcommand\ChangeRT[1]{\noalign{\hrule height #1}}

\usepackage{floatrow}
\newcommand{\bbR}{\mathbb {R}}

\newcommand{\cQ}{\mathcal{Q}}
\newcommand{\cG}{\mathcal{G}}

\newcommand{\cD}{\mathcal{D}}

\renewcommand\footnoterule{\vspace*{0.3cm}\hrule width 2.5cm\vspace*{0.2cm}}

\graphicspath{{./figures/}}

\begin{document}

\title{Frame invariance and scalability of neural operators for partial differential equations}

\author{Muhammad I. Zafar\affil{1}, Jiequn Han\affil{2}\comma\corrauth, Xu-Hui Zhou\affil{1}, and Heng Xiao\affil{1}}

\emails{{\tt jiequnhan@gmail.com} (Jiequn Han), {\tt hengxiao@vt.edu} (Heng Xiao)}

\address{\affilnum{1}\ Kevin T. Crofton Department of Aerospace and Ocean Engineering, Virginia Tech, Blacksburg, Virginia, USA\\
\affilnum{2}\ Center for Computational Mathematics, Flatiron Institute, New York, USA }

\begin{abstract}
Partial differential equations (PDEs) play a dominant role in the mathematical modeling of many complex dynamical processes. Solving these PDEs often requires prohibitively high computational costs, especially when multiple evaluations must be made for different parameters or conditions. After training, neural operators can provide PDEs solutions significantly faster than traditional PDE solvers. In this work, invariance properties and computational complexity of two neural operators are examined for transport PDE of a scalar quantity. Neural operator based on graph kernel network (GKN) operates on graph-structured data to incorporate nonlocal dependencies. Here we propose a modified formulation of GKN to achieve frame invariance. Vector cloud neural network (VCNN) is an alternate neural operator with embedded frame invariance which operates on point cloud data. GKN-based neural operator demonstrates slightly better predictive performance compared to VCNN. However, GKN requires an excessively high computational cost that increases quadratically with the increasing number of discretized objects as compared to a linear increase for VCNN.
\end{abstract}


\maketitle

\section{Introduction}
\label{sec:introduction}

A wide class of important engineering and physical problems are governed by partial differential equations (PDEs) describing the conservation laws. Extensive research efforts have gone into formulating and solving these governing PDEs. Despite significant progress, major challenges remain related to the computational costs of solving complex PDEs for real life problems like turbulent flows, laminar-turbulence transition and climate modeling. To avoid these prohibitively high computational costs, developing accurate and efficient numerical approximations or surrogate models for PDEs has been a key area of research~\cite{quarteroni2014, lucia2004, peherstorfer2016, taira2020modal}. Machine learning based models~\cite{lu2019deeponet,wang2021DeepONets,li2021fourier,zhou2021frame,kovachki2021neural} have the potential to provide a significantly faster alternatives to the traditional methods~\cite{zienkiewicz1977fem, rozza2008} of surrogate modeling. For accuracy and physical realizability, it is desired that these machine learning based models closely mimic the properties of the governing PDEs. 

One of the key features of these PDEs is \emph{frame invariance}, which is an intrinsic property of all equations in classical mechanics from Newton's second law to Navier Stokes equations. It signifies that the behavior of the physical systems does not depend on the origin or orientation of the reference frame of the observer. A model is frame-invariant to a transformation if the transformation of the input data does not alter the output of the function or model. In the context of fluid mechanics, any scalar variable like pressure or velocity magnitude is independent of any translation or rotation of the reference frame. 
Specifically, for example, a vector-to-scalar constitutive mapping $f: \bm{q} \mapsto \tau$ should remain unchanged in the frame rotated by matrix $\mathsf{R}$, i.e., the same mapping should be valid for $f: \mathsf{R} \bm{q} \mapsto \tau$. In other words, for any mapping $f: \bm{q} \mapsto \tau$  to be frame-invariant, the mapping $f: \bm{q}' \mapsto \tau$ should hold for any rotation matrix~$\mathsf{R}$, with $\bm{q}'\equiv \mathsf{R}\bm{q}$ being the input vector in the new, rotated coordinate system. Invariance with respect to other transformations (e.g., translation of origin or change of reference velocity) can be defined and interpreted similarly. For solid mechanics, the magnitude of the deformation tensor is invariant to the orientation of the reference frame or the origin of the frame -- it is an objective quantity regardless of the observer. Clearly, any modeling equations or constitutive relations should faithfully reflect such invariance or symmetries.
Furthermore, in the numerical solutions of these PDEs, frame invariance also requires permutational invariance, which ensures independence of the results from the order in which the discretized objects are indexed. Examples of such objects include elements, cells, grid points, or particles, depending on the specific numerical method used.
In a hypothetical simple example, assume the scalar $\tau$ at a given location $\bm{x}_0$ is a function of the vectors $\bm{q}_1$, $\bm{q}_2$, and $\bm{q}_3$ at three cells in the neighborhood of $\bm{x}_0$. The output $\tau$ must remain identical whether the input is $(\bm{q}_1, \bm{q}_2, \bm{q}_3)$, $(\bm{q}_3, \bm{q}_2, \bm{q}_1)$, or any other ordering of three vectors. In other words, permutation invariance demands that $\tau$ must only depend on the set as a whole and not on the ordering of its elements.
Embedding these invariance properties in machine learning based models can significantly improve the generalizability of these models~\cite{ling2016jcp, han2021machine, kaandorp2020invariance, frezat_2021, wang2021incorporating}.

In this work, we are particularly interested in using neural network-based solution operators, called~\emph{neural operators}~\cite{lu2019deeponet,kovachki2021neural}, to solve PDEs with a guarantee of frame invariance. Mathematically, given a PDE system, the \emph{solution operator} maps the input function such as the initial condition, boundary condition, or auxiliary field to the output PDE solution. The solution operator plays an essential role in many applications like surrogate modeling, constitutive modeling, and Bayesian inverse problem. Note that the solution operator is defined with respect to a family of PDEs with different input functions, in contrast to classical methods like finite difference methods and modern machine learning based methods like physics-informed neural network~\cite{raissi2019pinn, gao2021pignn} that aim to determine the solution for a single instance of the PDE. Those methods are computationally expensive when multiple evaluations are required. In contrast, neural operators aim to approximate the solution operator through an explicit mapping starting from the input function so that it operates significantly faster. It is also designed to be mesh-independent. Of course, learning a solution operator is significantly more challenging than determining the solution to a PDE in a single instance. But once the network parameters are properly optimized, the neural operator can predict the solution of a new instance of the PDE fast. In this work we examine two such neural operators, graph kernel networks and vector cloud neural networks.

Recently, graph based discretization of the physical domain has found prominence for the development of neural operators and more broadly, scientific simulations~\cite{pfaff2021learning}, using graph neural networks~\cite{scarselli2009gnn}.
Graph kernel networks~\cite{anandkumar2020neural,li2020multipole} is a specialized class of graph neural networks for implementing nerual operators. GKN operates on graph-structured data, which is consistent with grids or data structures used in computational methods.
A kernel network is used to incorporate nonlocal dependencies from other graph nodes. GKN can exhibit translational invariance if relative spatial information is considered for neighbouring nodes in the graph. In that case global attribute of the graph is invariant to a translation of the graph-structured data in the spatial domain. Similarly, the global arithmetic operation to determine the global attribute of the graph-structured data embeds permutational invariance, which guarantees independence from the order in which graph nodes are indexed. Rotational invariance can be achieved by alternative methods for graph networks. A simpler method, borrowed from computer vision, is to augment the training data with differently rotated coordinate systems. However, such a method is inefficient as it increases the data size and training time (by approximately a factor of over ten as observed in~\cite{ling2016jcp}). Moreover, the data augmentation does not guarantee frame invariance strictly -- rather, the trained network only has approximate rotational invariance~\cite{lyle2020benefits}. Another method to achieve rotational invariance is to use rotational invariant input features. This work will discuss such a method that can be effectively implemented with GKN to achieve frame invariance in a data-efficient manner. 

The computational complexity of GKN is dependent on the sparsity of the underlying graph structure. When the graph is complete, i.e., every pair of nodes is connected with an edge, the computation cost scales quadratically with the number of nodes or data points in the graph. For some scientific problems, the graph size for capturing nonlocal dependencies can be of the order of hundreds or thousands nodes, and the graph is densely connected, like in the case of high Reynolds number flows. In such cases, GKN does not scale well and can become prohibitively expensive. This computational disadvantage of GKN was also discussed in~\cite{anandkumar2020neural} where a graph kernel network was used to emulate the Green's function for the Poisson equation. 

Vector cloud neural network (VCNN) provides an alternative strategy to develop a neural operator with embedding frame invariance~\cite{zhou2021frame}.
The nonlocal information of the PDE is represented by a \emph{vector cloud}, an arbitrarily arranged group of points akin to graph-structured data, with each point (node) having a feature vector attached to it. Input matrix $\mathcal{Q}$ is organised by stacking feature vectors corresponding to every point in the cloud. This vector cloud is mapped to a global scalar attribute of the cloud, through a frame-invariant neural network-based model. Spatial coordinates for neighbouring points in the cloud are organised relative to the central point of the cloud, which ensures translational invariance. Pairwise projection $\mathcal{D}^\prime = \mathcal{Q} \mathcal{Q}^{\top}$ of feature vectors is used to obtain rotational invariant features. 
Furthermore, an embedding network is used to obtain a learned basis $\mathcal{G}$ and the rotational invariant features are projected on this learned basis as $\mathcal{D}=\frac{1}{n^2}\mathcal{G}^{\top} \mathcal{D}^\prime \mathcal{G}$ where $n$ is the number of points in the cloud. Such projections output the average point features on the learned basis and guarantee permutational invariance in the model. Invariant feature matrix $\mathcal{D}$ is then nonlinearly mapped to the unknown scalar variable of PDE through a fitting network.
For predicting the unknown variable in the entire field, VCNN has shown to be computationally more efficient compared to traditional mesh-based PDE solvers such as those in finite volume methods as the number of mesh cells in the computational domain scales~\cite{zhou2021frame}.
Another comparable network architecture with embedded frame invariance, PointNet~\cite{qi2017pointnet}, also operates on point clouds to learn point-wise latent features by mapping through a shared embedding network. A global feature is then extracted by using an aggregate function (like max-pooling) to ensure permutational invariance. Rotational invariance is ensured by the use of scalar input features only or together with vector features defined in a rotational equivariant coordinate system~\cite{xiao2019SRINet}.
Considering grid vertices of a CFD domain as a point cloud, PointNet has been used to define an end-to-end mapping from the spatial positions of the grid vertices over irregular geometries to the corresponding velocity and pressure fields of the flow~\cite{kashefi2021pointnet}.

In the present work, we use flows over a symmetric airfoil to investigate the frame invariance property and scalability of the neural operators discussed above. As a proof of concept, we perform this investigation for a neural operator developed for a transport equation of a scalar variable. The behavior of such a scalar variable can also be closely related to scalar quantities in turbulence modeling including turbulent kinetic energy (TKE) and turbulent length scale~\cite{wilcox06turbulence}. In this preliminary work, we consider a hypothetical dimensionless concentration tracer $\tau$, which is transported by a laminar velocity field $\bm{u}$. Although the originally proposed GKN does not ensure rotational invariance, we have modified it to use rotational invariant input features and map them to the scalar variable $\tau$. Such results are then compared against the embedded frame invariance of VCNN. The results are analyzed for accuracy and computational efficiency.

The rest of the manuscript is organised as follows. Methodologies used for this work are presented in Section~\ref{sec:method}. Generation of data for training and testing of the neural operators is discussed. Both neural operators, GKN and VCNN, are also presented while elaborating on their frame invariance properties. Section~\ref{sec:results} presents the analysis of frame invariance properties, predictive performance and computational efficiency of both models. Section~\ref{sec:conclusion} concludes the paper.

\section{Methodology}
\label{sec:method}

\subsection{Generation of data for training and testing}
Flow over a symmetric NACA-0012 airfoil in a two-dimensional space is considered for this study. The airfoil contour was obtained from a public domain source, UIUC Airfoil Coordinates Database \cite{uiuc}. C-mesh flow domain is used to appropriately distribute the mesh density along the critical areas of flow around an airfoil. For an airfoil with unit chord length, the width of the flow domain extends 5 times the chord length towards the upstream direction and 10 times the chord length towards the downstream direction of the airfoil leading edge. The domain extends 5 times the chord length in the direction normal to the airfoil chord. 

The scalar quantity $\tau$ is a hypothetical, dimensionless concentration tracer, which is transported by the velocity field $\bm{u}$. A simplification is assumed for this study that the velocity field is not affected by the concentration tracer. To compute the scalar field $\tau(\bm{x})$, the transport PDE is solved with a steady, laminar flow field $\bm{u}(\bm{x})$. Resembling the transport equation of other scalar quantities like turbulent kinetic energy, eddy viscosity, etc., the transport PDE for concentration $\tau$ is given as: 
\begin{equation}
    \label{eq:transport_eqn}
    \bm{u} \cdot \nabla \tau-\nabla \cdot\left(C_{\nu} \nabla \tau\right)=\mathsf{P}-\mathsf{E}
\end{equation}
where $\mathsf{P}$ indicates the production term, $\mathsf{E}$ indicates the dissipation term and $C_\nu$ represents the diffusion coefficient. Both the production and dissipation terms depend on the scalar quantity $\tau$. The production is further dependent on the mixing length $\ell_m$ and strain-rate magnitude $\|\mathsf{s}\|$. Both terms are given as:
\begin{align*}
    & \mathsf{P}=C_{g} \ell_{m} \sqrt{\tau} s^{2} \text { and } \mathsf{E}=C_{\zeta} \tau^{2} \\
    \text { with } \quad &s=\|\mathsf{s}\|=\left\|\nabla \bm{u}+(\nabla \bm{u})^{\top}\right\|
\end{align*}
where $C_g$ and $C_{\zeta}$ are production and dissipation coefficients, respectively. Further details for this transport equation and solver have been discussed in detail by~\cite{zhou2021frame}.

\begin{figure}[!htb]
    \centering
    \subfloat[Reference frame for the training dataset at AOA = 10\degree]{
    \includegraphics[width=0.37\textwidth]{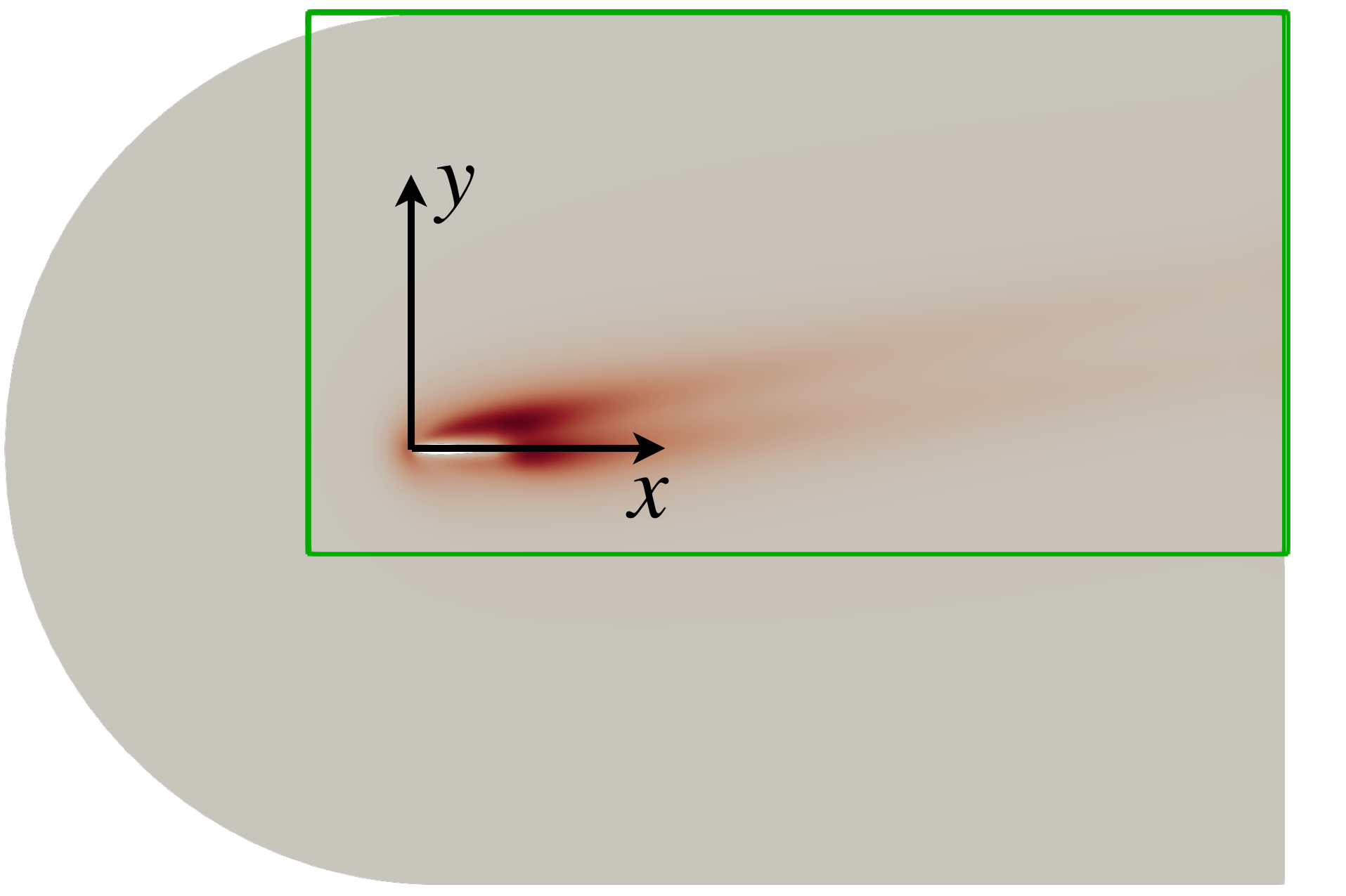}}
    \qquad
    \subfloat[Transformed reference frame for the testing dataset at AOA = 15\degree]{
    \includegraphics[width=0.45\textwidth]{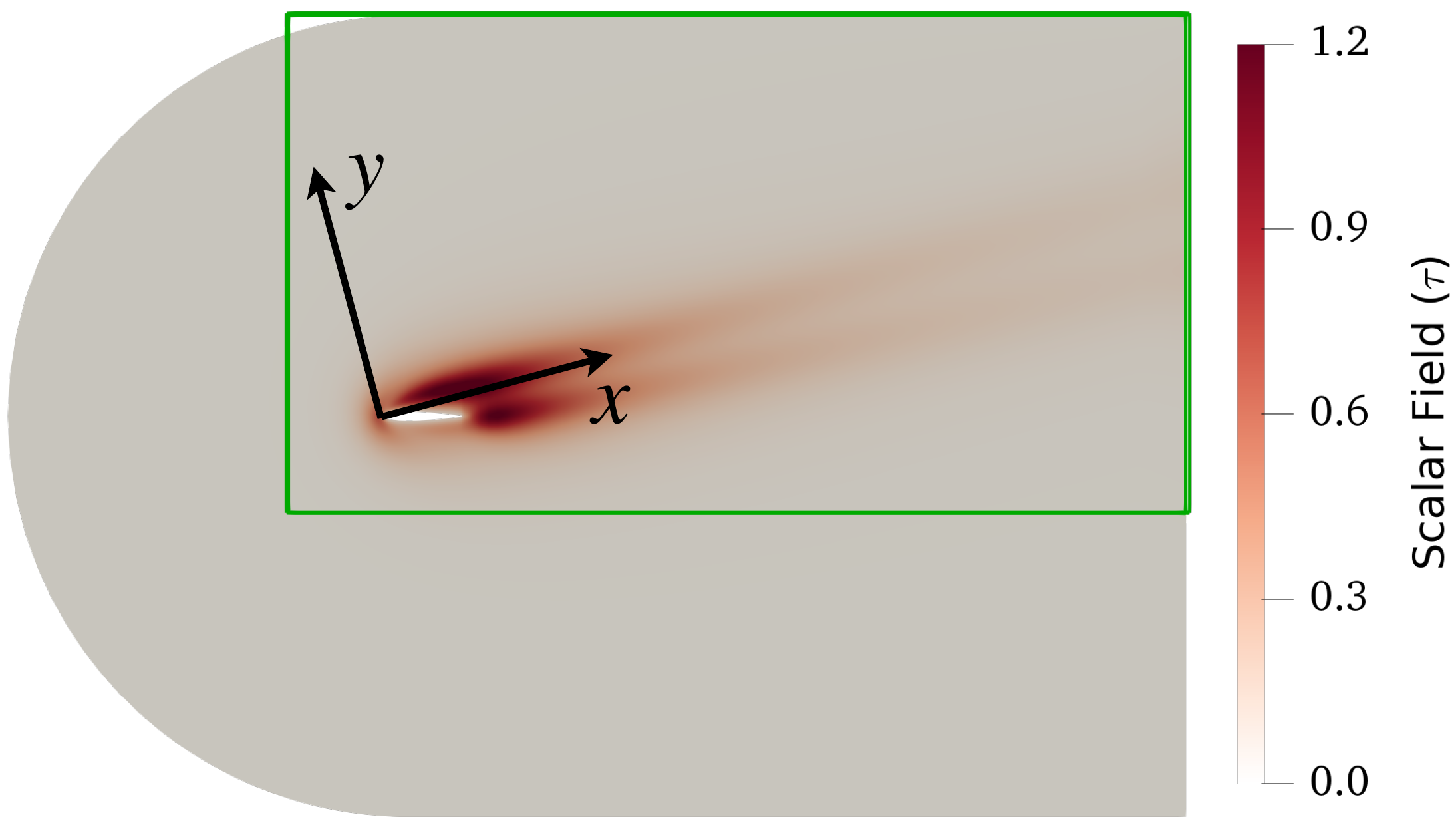}}
    \caption{Different frames of reference used for training and testing datasets. Contours show the scalar field $\tau$ transported by flow over a symmetric airfoil (NACA-0012). The training dataset is based on Cartesian coordinate system with origin coinciding the airfoil leading edge and $x$-axis aligned with the airfoil chord. Coordinate system for the testing dataset is transformed/rotated by the given angle of attack (AOA), with origin still coinciding with the airfoil leading edge. Red box indicates the truncated flow domain. Data outside the green box is discarded to avoid redundancy.}
    \label{fig:train_test_data}
\end{figure}

Laminar flow field is simulated first to obtain the velocity field $\bm{u}$. Using this velocity field, scalar field $\tau$ is then computed by solving transport PDE for concentration tracer. Data has been generated for a set of angles of attack (AOA) and divided into training and testing dataset as:
\begin{align*}
    & \text{Training dataset} : [10\degree,\ 20\degree,\ 30\degree], \\
    & \text{Testing dataset}: \enskip [5\degree,\ 15\degree,\ 25\degree,\ 35\degree].
\end{align*}
Testing dataset has been organised to interpolate to testing data at AOA = $(15\degree,\ 25\degree)$ and extrapolate to testing data at AOA = $(5\degree,\ 35\degree)$ flow conditions. Furthermore, in the testing dataset, a transformed reference frame is used, which is rotated by a value equal to the AOA in the corresponding flow. Such distinction is demonstrated in Fig.~\ref{fig:train_test_data}. This transformation of the reference frame is specifically implemented to analyze the frame invariance behavior of the neural operators. Furthermore, to avoid redundancy, the data is extracted from a truncated domain by discarding the region which does not have any concentration tracer distribution at any specified flow conditions.  

To generate data for nonlocal mapping from a patch of flow field $\bm{u}(\bm{x})$ to concentration $\tau$ at the point of interest, the setting is largely adapted from \cite{zhou2020nonlocal,zhou2021frame}. The flow features' vector $\bm{q} = \llbracket \bm{x}^{\prime}, \bm{u}, \bm{c} \rrbracket \in \mathbb{R}^{11}$ used at each point is chosen to include relative coordinate features $\boldsymbol{x}^{\prime}=\left[x^{\prime}, y^{\prime}\right]^{\top}$, flow velocity $\bm{u}$ and seven additional scalar quantities, collectively denoted by $\bm{c}$. Note that our convention in this paper is that all vectors are column vectors, and $\llbracket\cdot \rrbracket$ denotes vertical concatenation. 
Those scalar quantities in $\bm{c}$ include mesh cell volume ($\theta$), magnitude of strain rate ($s$), boundary cell indicator ($b$), velocity magnitude ($\mathrm{u}$), wall distance ($\eta$), proximity of cloud center ($r$), proximity in local velocity frame ($r^\prime$). Detailed descriptions and mathematical definitions of these scalar quantities can be found in~\cite{zhou2021frame}.

\begin{figure}[!htb]
    \centering
    \includegraphics[width=0.7\textwidth]{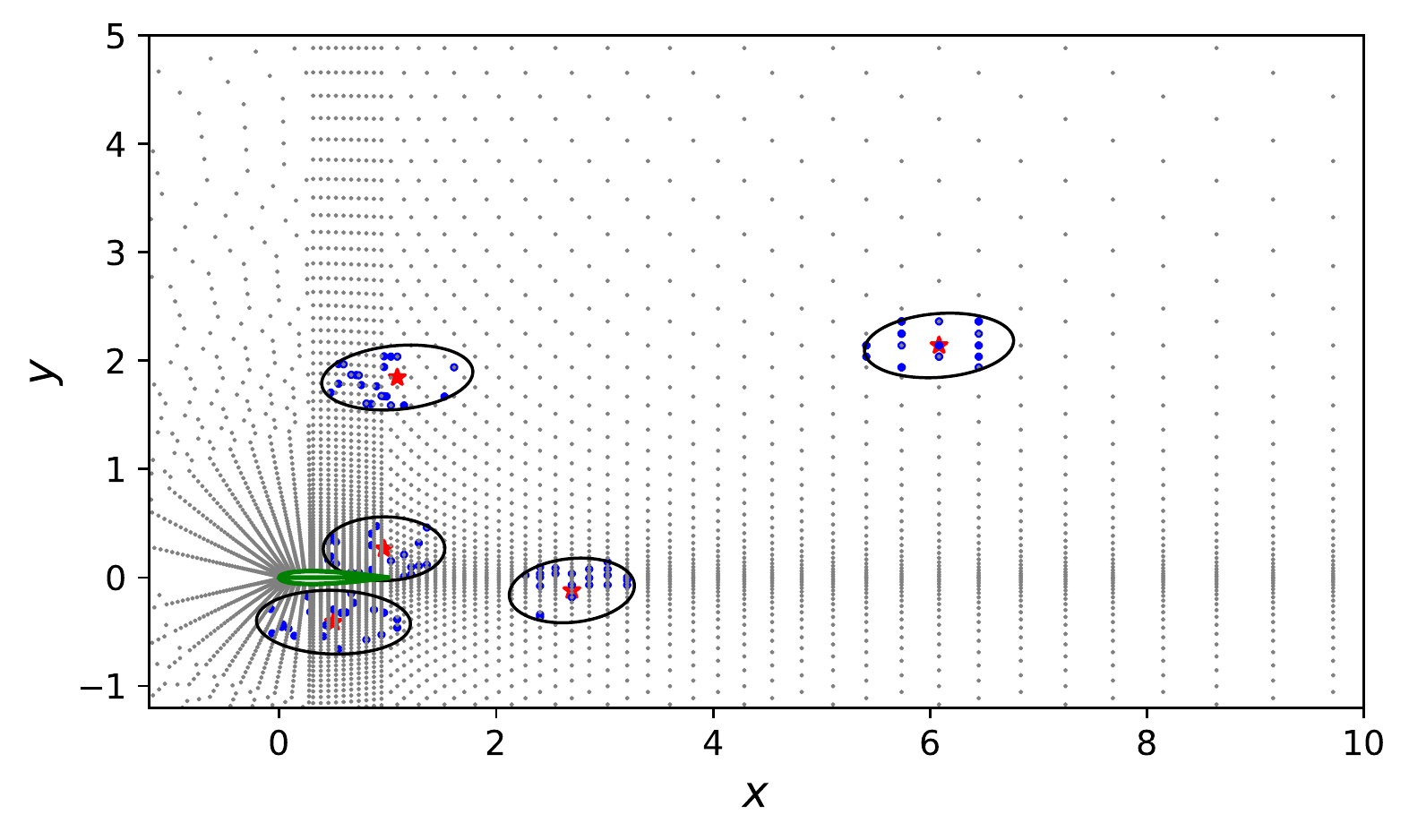}
    \caption{Sampling a cloud of data points for neural operator mapping. Airfoil surface ($\color{teal} \bullet$) is shown in green color. The gray dots ($\color{gray!80} \bullet$) indicate cell centers, showing every seventh cell with respect to the index of the structured mesh for clarity. For any location ($\color{red} \star$) where the concentration $\tau$ is to be predicted, ellipse (\protect\tikz{ \protect\draw (0,-5) ellipse (6pt and 2.5pt);}) shows the physical region of influence as determined by the velocity, diffusion and dissipation coefficients at that point; see Eq~\eqref{eq:influence_region}. The blue dots ($\color{blue} \bullet$) are the randomly sampled cell centers within the region of influence and the feature vectors attached to them are used as input matrix $\cQ$ to predict~$\tau$.}
    \label{fig:stencil}
\end{figure}

Flow features are considered at randomly selected fixed number of points within the region of influence around the point of interest where concentration $\tau$ has to be determined. Fig.~\ref{fig:stencil} illustrates the regions of influence and respective sampled data points corresponding to different points of interest in the flow domain. The extent of the region of influence, represented by an ellipse, depends on the velocity, diffusion and dissipation coefficients at the point of interest. The half-lengths $\ell_1$ and $\ell_2$ of the major and minor axis of the ellipse are determined based on the specified relative error tolerance $\epsilon$:
\begin{equation}
\label{eq:influence_region}
    \ell_{1}=\left|\frac{2 \nu \log \epsilon}{\sqrt{|\bm{u}|^{2}+4 \nu \zeta}-|\bm{u}|}\right| \quad \text { and } \quad \ell_{2}=\left|\sqrt{\frac{\nu}{\zeta}} \log \epsilon\right|
\end{equation}
where $\nu$ and $\zeta$ are the diffusion and dissipation coefficients, for which details can be found in \cite{zhou2020nonlocal}. The major axis of the ellipse aligns with the direction of the velocity $\bm{u}$. Each input-output combination for the neural operators then corresponds to an elliptical patch of flow domain, where input data is the stack of feature vectors arranged as input feature matrix $\mathcal{Q} = [\bm{q}_1, \bm{q}_2, \cdots, \bm{q}_n]^{\top}$ and output is the scalar variable $\tau$ at the point of interest in that elliptical patch.
Stencil size ($n$) represents the number of randomly sampled fixed number of cloud data points from that elliptical patch, or equivalently the amount of information considered from the region of nonlocal influence. Constant stencil size is considered throughout the flow domain. For the locations where the number of data points in the ellipse is smaller than the specified stencil size, the available data points are repeatedly sampled. 

\subsection{Graph Kernel Network}
Graph kernel network is based on graph neural networks, a class of neural networks adapted to leverage the structure and properties of graphs. A graph is an extremely powerful and general representation of data used to express a collection of entities (nodes) and the connections (edges) between them. Depending on the application, a graph can have attributes/features attached to the nodes and/or edges, and in some cases each graph can have a global attribute. Such specialized neural networks have found several practical applications such as antibiotic discovery~\cite{stokes2020antibiotic}, physics simulations~\cite{gonzalez2020complexphysics}, fake news identification~\cite{monti2019FakeND}, traffic prediction~\cite{jiang2021} and recommendation systems~\cite{eksombatchai2018}. 

In the context of PDEs, graph nodes can be associated with the mesh cells and the cells can be connected through edges. To predict $\tau$ at any particular point, one can construct a graph among the neighbouring nodes in the region of influence and assimilate the nonlocal information through corresponding edges. Scalar quantity $\tau$ is the global attribute for each graph which can then be determined by considering the attributes of all the nodes in the graph.

\begin{figure}[!htb]
    \includegraphics[width=0.99\textwidth]{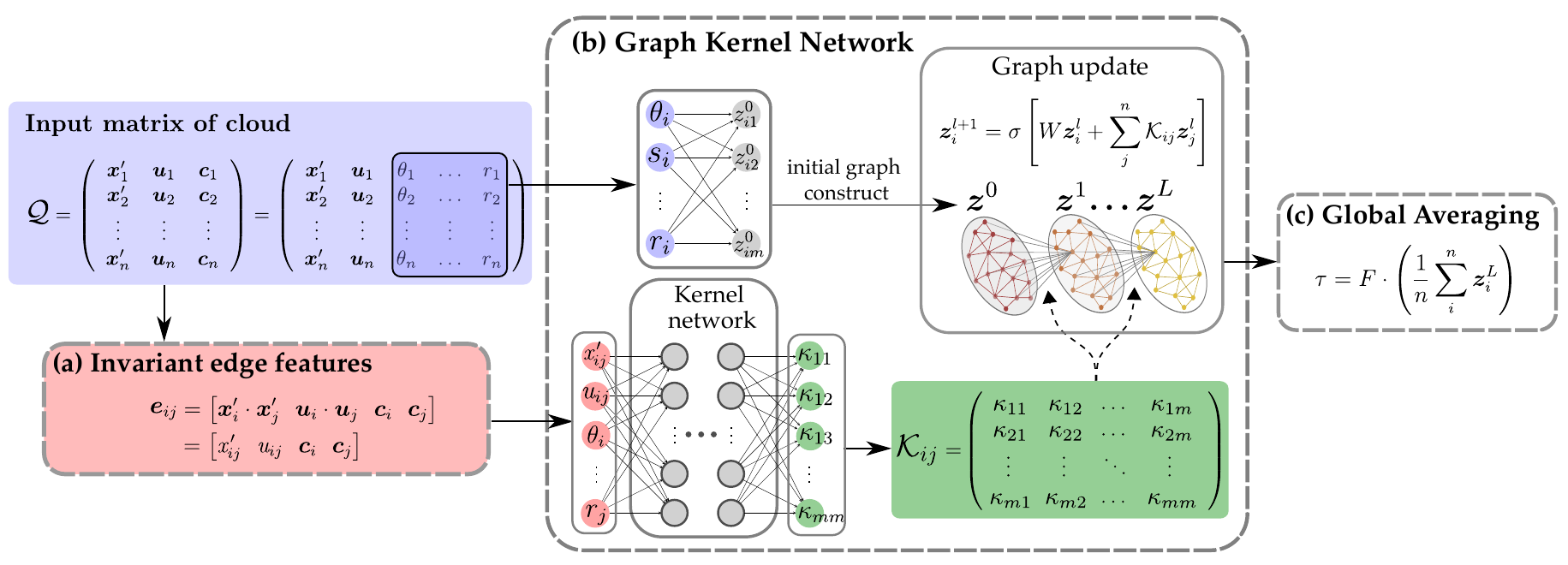}
    \caption{Detailed schematic of graph kernel network for region-to-point mapping from input matrix $\cQ$ to scalar variable $\tau$: (a) organises rotational invariant edge features between two points/nodes in the cloud; (b) map scalar features from input matrix to initial graph-based representation $\bm{z}^0$ in latent space; then update this graph by assimilating nonlocal information in invariant edge features ${\bm{e}}_{ij}={\bm{e}}(\bm{q}_i, \bm{q}_j)$ through  neural network based kernel function $\mathcal{K}_{ij} = \mathcal{K}({\bm{e}}_{ij})$; (c) perform the averaging operation over $\bm{z}_i^L$ in the last GKN layer and the inner product with learnable parameters' vector to determine the global attribute of the graph scalar variable $\tau$. $W \in \mathbb{R}^{m \times m}$ and $F \in \mathbb{R}^{1 \times m}$ are matrices of learnable parameters.}
    \label{fig:schematic_gkn}
\end{figure}

Li et~al.~\cite{anandkumar2020neural} proposed GKN to learn mappings between function spaces represented by graph-based data. The mapping is formulated by using a neural network-based kernel function. Fig.~\ref{fig:schematic_gkn} 
shows a schematic of the GKN architecture in the context of our example. 
The graph is completely connected, and the connection between two nodes in a graph is attributed by edge features as:
\begin{equation}
    \Hat{\bm{e}}(\bm{q}_i, \bm{q}_j) = \left[ \bm{x}_i^\prime, \quad \bm{x}_j^\prime, \quad \bm{u}_i, \quad \bm{u}_j, \quad \bm{c}_i, \quad \bm{c}_j \right],
    \label{eq:nonRI_input}
\end{equation}
which are obtained by \emph{concatenating} the flow feature vectors $\bm{q} = \llbracket \bm{x}^{\prime}, \bm{u}, \bm{c} \rrbracket$ of the corresponding nodes. 
Here $i$ and $j$ represent two corresponding nodes of the edge. 
Unlike scalar features $\bm{c}$, relative coordinates $\bm{x}^\prime$ and velocity $\bm{u}$ does not possess rotational invariance, rendering edge features in Eq.~\eqref{eq:nonRI_input} dependent on the choice of frame orientation. Therefore, to ensure rotational invariance, we propose here to use rotational invariant edge features by \emph{projecting the vectors} on to each other and by \emph{concatenating the scalars}, which are specifically given as:
\begin{equation}
    \bm{e}(\bm{q}_i, \bm{q}_j) = \left[ \bm{x}_i^{\prime\top} \bm{x}_j^\prime, \quad \bm{u}_i^{\top}  \bm{u}_j, \quad \bm{c}_i, \quad \bm{c}_j \right].
    \label{eq:ri_input}
\end{equation}
Note that except the difference between Eq.~\eqref{eq:nonRI_input} and Eq.~\eqref{eq:ri_input} in edge features, our modified GKN has the same structure as that in~\cite{anandkumar2020neural}, as described below.

Given edge features, kernel integration is computed to approximate mapping from one graph-based representation to another. Each graph-based representation is referenced as a layer ($l$) and multi-layer GKN enables capturing long-range correlations in graph-based data~\cite{anandkumar2020neural}. Initial graph-based representation in latent space was proposed in \cite{anandkumar2020neural} to be based on the complete feature vector $\bm{q}$ on each node. However, here we consider only the scalar features $\bm{c}$ for initialization, which possesses rotational invariance. Each GKN layer update, referred to as graph convolution, can then be expressed as:
\begin{align}
    \bm{z}^0_i = & \ Z\bm{c}_i, \\
    \bm{z}^{l+1}_i = & \ \sigma \left[ W \bm{z}^{l}_i \ + \ \frac{1}{n} \sum_{j=1}^n \mathcal{K}({\bm{e}}_{ij}) \bm{z}^{l}_j \right], \quad \quad \quad l = 1,2 \cdots L-1,
\end{align}
where $i = 1, 2 \ldots n$ indexes the nodes in the graph representation, $\bm{z}^0_i$ is a vector of $m$-dimension representing the latent node features of $i$th node in the initial graph representation,
${\bm{e}}_{ij}$ is the shorthand notation of ${\bm{e}}(\bm{q}_i, \bm{q}_j)$,
$Z \in \mathbb{R}^{7 \times m}$ represents the learnable parameters of a fully connected linear layer, $W \in \mathbb{R}^{m \times m}$ represents a matrix of learnable parameters for information passing among nodes, $L$ is the total number of layers or depth of GKN and $\mathcal{K}$ represents the neural network based kernel function. Details of the kernel network architecture have been mentioned in the Appendix~\ref{architecture} (Table~\ref{tab:gkn-details}). The output of $\mathcal{K}$ is $m^2$-dimensional and reshaped to a square matrix. Latent features of each node in the graph representation ($\bm{z}^l_i$) are updated using kernel network to incorporate the information from all the connected nodes and learnable parameters' matrix $W$ to process the latent features of the node itself in the previous layer. Hence each node in the updated graph representation is connected to every node in the previous graph representation. This graph update is performed for specified number of times enumerated as the depth ($L$) of GKN. 

For scalar field $\tau$ over an airfoil, GKN is purposed for region-to-point mapping. In this regard, to obtain a global attribute for the graph representation at layer $L$, a global averaging operation is performed as:
\begin{equation}
    \tau=F \cdot\left(\frac{1}{n} \sum_{i=1}^{n} \bm{z}_{i}^{L}\right)
\end{equation}
where $F \in \mathbb{R}^{1 \times m}$ represents learnable parameters and $\bm{z}^L_i$ represents the latent features of $i$-th node in the last GKN layer ($L$). This global averaging operation also guarantees permutational invariance for GKN. With translational invariance already guaranteed, rotational input features and global averaging operation ensures the frame invariance property of the GKN based model presented here.

\subsection{Vector Cloud Neural Network}

Zhou et~al.~\cite{zhou2021frame} recently proposed a frame-independent neural operator, which is able to map to a  scalar variable at a point based on the nonlocal information in its neighbourhood. Similar to graph neural networks, such nonlocal information is taken into account as an arbitrarily arranged group of points referred to as \emph{vector cloud}, with a feature vector attached to each point. Since this nonlocal information can be captured from arbitrarily arranged points, irrespective of the ordering of points in the cloud, VCNN is a powerful tool to deal with unstructured meshes in scientific simulations.

\begin{figure}[!htb]
    \includegraphics[width=0.99\textwidth]{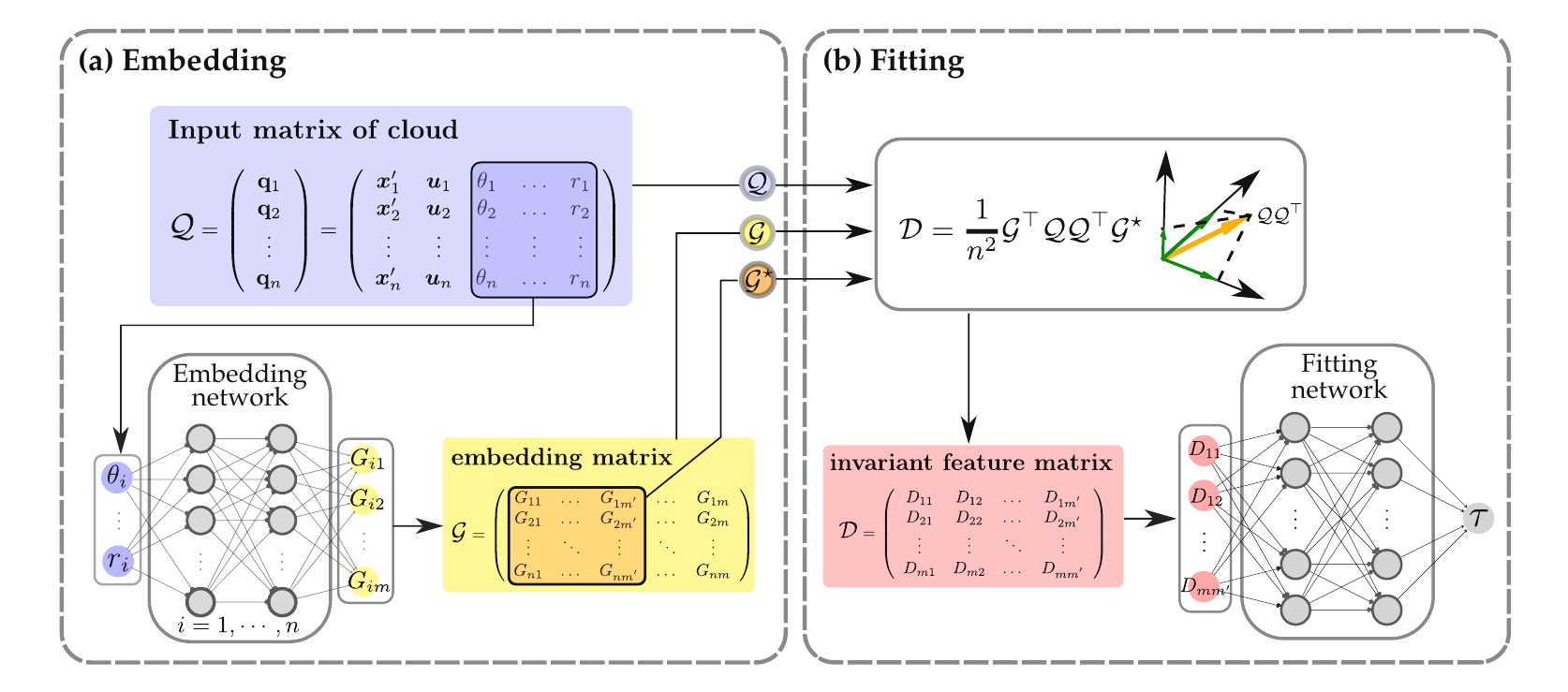}
    \caption{Detailed schematic of the vector cloud neural network for region-to-point mapping from input matrix $\cQ$ to scalar variable $\tau$: (a) map the scalar features in $\cQ$ to a set of learned basis $\cG$ through an embedding network, $\cG^{\star} \in \mathbb{R}^{n \times m^{\prime}}$ is the first $m^{\prime}(\leq m)$ columns of $\cG$; (b) project the pairwise inner-product matrix $\mathcal{Q} \mathcal{Q}^{\top}$ to the learned embedding matrix and its submatrix to obtain invariant feature matrix $\mathcal{D} \in \mathbb{R}^{m \times m^{\prime}}$; which is then mapped through fitting network to predict scalar variable $\tau$.}
    \label{fig:schematic_vcnn}
\end{figure}

Region-to-point mapping from input matrix $\mathcal{Q}$ of vector cloud features to the scalar variable $\tau$ is illustrated in Fig.~\ref{fig:schematic_vcnn} as a schematic figure of VCNN. Input matrix $\mathcal{Q}=[\bm{q}_1, \cdots, \bm{q}_n]^\top \in \mathbb{R}^{n\times 11}$ is organised by stacking feature vectors corresponding to every point in the cloud. Each row in $\mathcal{Q}$ is a feature vector on a point.
With the usage of relative coordinates $\bm{x}^\prime$ and pairwise projections $\mathcal{Q} \mathcal{Q}^{\top}$ among the feature vectors, a direct mapping of the form $\tau=\hat{g}\left(\mathcal{Q} \mathcal{Q}^{\top}\right)$ can ensure translational and rotational invariance. However to guarantee frame invariance completely, permutational invariance also needs to be embedded. For such purpose, an embedding network is used, as shown in Fig.~\ref{fig:schematic_vcnn}(a). Note that scalar features $\bm{c}_i$ already possess translational and rotational invariance. These scalar features $\bm{c}_i$ are mapped through an embedding network to obtain $m$-dimensional embedded weights $\phi(\bm{c}_i)$ corresponding to each point, and form an embedding matrix $\mathcal{G}=[\phi(\bm{c}_1),\cdots, \phi(\bm{c}_n)]\in \mathbb{R}^{n \times m}$ corresponding to all the points in the cloud. To introduce permutational invariance, an order-removing transformation is performed as $\mathcal{G}^{\top} \mathcal{Q}$. Combining it with the pairwise projections $\mathcal{Q} \mathcal{Q}^{\top}$ among the feature vectors, invariant feature matrix can be obtained as 
\begin{equation}
\mathcal{D}=\frac{1}{n^{2}} \mathcal{G}^{\top} \mathcal{Q} \mathcal{Q}^{\top} \mathcal{G}^{\star},
\label{eq:D}
\end{equation}
where $\mathcal{G}^{\star}$ is taken as a subset of $\mathcal{G}$ to save computational cost. Equivalently, this can be written as $\mathcal{D} = \mathcal{L}  \mathcal{L}^{\star\top}$ where
\begin{equation}
    \mathcal{L} =  \frac{1}{n} \mathcal{G}^{\top} \mathcal{Q}
    \label{eq:L}
\end{equation}
and $\mathcal{L}^\star$ similarly defined for $\mathcal{G}^\star$.  The normalization by the number $n$ of points in the cloud allows the training and prediction to use different number of sampled point in the cloud. The obtained invariant feature matrix is then mapped through a fitting network to the scalar variable $\tau$, as shown in Fig.~\ref{fig:schematic_vcnn}(b).

The combination of embedding network, projection mapping and fitting network thus forms a complete VCNN model that has embedded frame invariance in the model. Detailed architectures of embedding and fitting networks have been mentioned in the Appendix~\ref{architecture} (Table~\ref{tab:vcnn-details}). Besides embedded frame invariance attribute, VCNN has demonstrated reasonable computational efficiency compared to traditional mesh-based PDE solvers such as those in finite volume methods for predicting the unknown variable in the entire field, especially as the number of mesh cells in the computational domain scales~\cite{zhou2021frame}.
With this perspective, the performance and scalability of VCNN to develop a frame-invariant neural operator is compared with that of GKN in the next section.

\subsection{Comparative analysis of GKN and VCNN}

\begin{figure}[!htb]
    \includegraphics[width=0.99\textwidth]{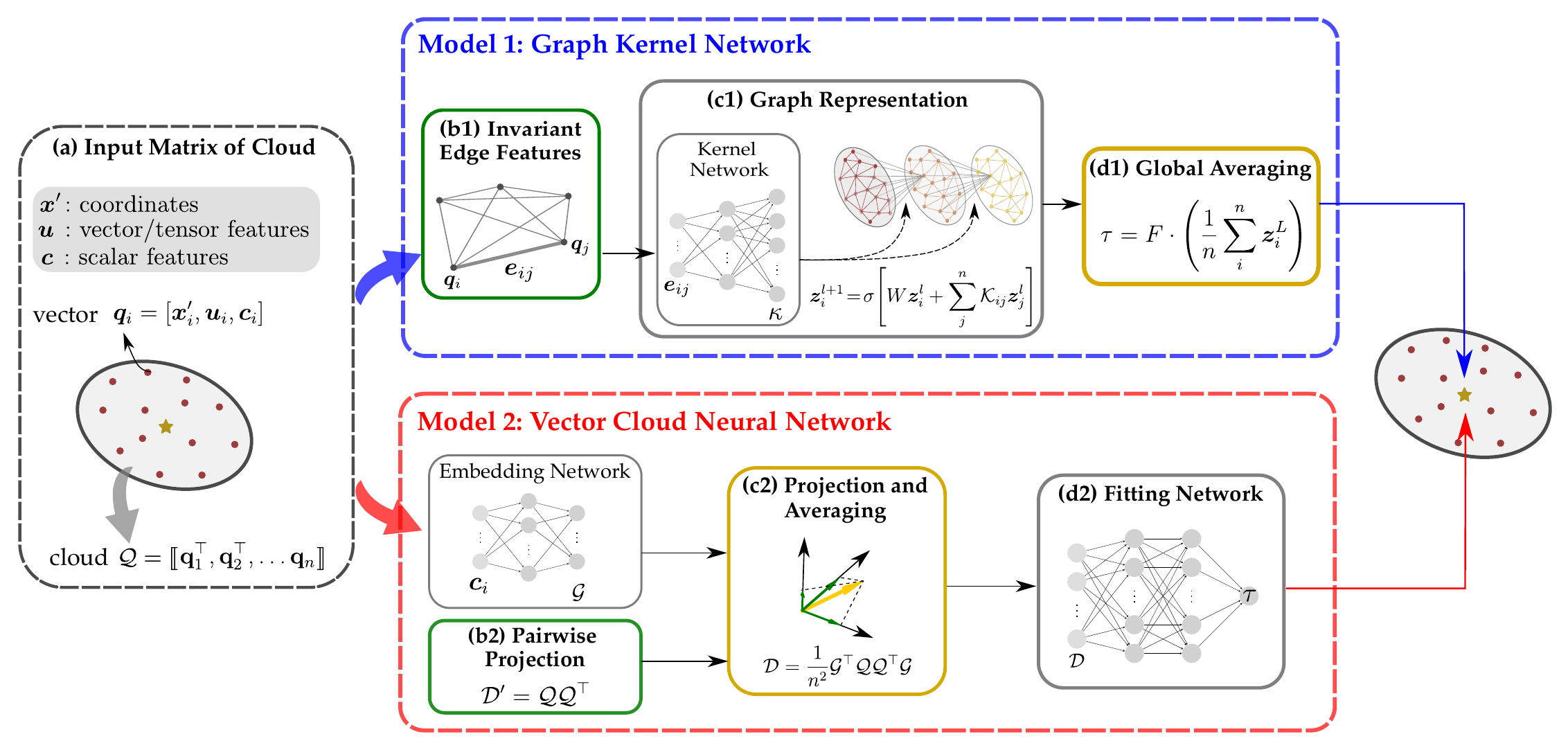}
    \caption{Schematic comparison of GKN and VCNN where different components of both models are outlined in the same color with respect to their role in achieving frame invariance for respective models. Translational invariance is ensured by the use of relative coordinates in the input cloud matrix $\mathcal{Q}$.  Components contributing to rotational invariance are outlined in green where rotational invariant edge features (b1) and pairwise projections $\mathcal{Q} \mathcal{Q}^{\top}$ (b2) among the feature vectors are used. Components contributing to permutational invariance are outlined in yellow where global averaging (d1) and order-removing transformation $\mathcal{D}=\frac{1}{n^2}\mathcal{G}^{\top} \mathcal{Q} \mathcal{Q}^{\top} \mathcal{G}$ in projection mapping (c2) are used. Nonlocal mapping is achieved by the use of kernel network (c1) and by the fitting network (d2).}
    \label{fig:schematic_gkn_vcnn}
\end{figure}

Both neural operators, GKN and VCNN, presented earlier are compared schematically in Fig.~\ref{fig:schematic_gkn_vcnn}. The same form of input matrix $\mathcal{Q}$ is provided for both models. Both models process the data and ensure frame invariance properties in different manners. Different components of each model can be related with respect to their role in ensuring frame invariance and accurately predicting the scalar variable.
To extract global nonlocal information, VCNN uses a fully-connected neural network (Fig.~\ref{fig:schematic_gkn_vcnn}(f)) for mapping from invariant feature matrix $\mathcal{D}$ to the scalar output $\tau$. In contrast, GKN uses a relatively intricate setting where a neural network based-kernel function incorporates the information from neighbouring points to update the latent graph representation sequentially over several layers, where the mapping between the layers is determined by a nonlinear neural network function $\mathcal{K}$ operating on the edges $\bm{e}(\bm{q}_i, \bm{q}_j)$. Such an elaborate process of incorporating nonlocal information provides GKN with more flexibility in learning the region-to-point mapping. However, the edge-based mapping is also the source of $\mathcal{O}(n^2)$ scaling of GKN in terms of training time and memory footprint, which will be shown later.  In contrast, the high dimension of the edge-based features (or equivalently pairwise project of node vectors) is reduced in the projection operation $\mathcal{L} = \frac{1}{n} \mathcal{G}^\top\mathcal{Q}$ in the embedding network of the VCNN. In fact, the pairwise project matrix $\mathcal{Q} \mathcal{Q}^\top$ in Eq.~\eqref{eq:D} is never formed explicitly in the implementation.
Another way to see the comparison between two models is by dropping the weights of both models and assuming one-layer graph convolution in GKN with edge features $\bm{q}_i \cdot \bm{q}_j$. In this case, if we denote the scalar feature matrix $\mathcal{C}=[\bm{c}_1, \cdots, \bm{c}_n]$, the feature extraction part of both models can be summarized as
\begin{align*}
    \text{GKN}: \quad   &\frac{1}{n^2}\mathcal{K}\left[\mathcal{Q}\mathcal{Q}^\top\right]\mathcal{C} + \frac{\mathcal{C}}{n},\\
    \text{VCNN}: \quad &\frac{1}{n^2}\sigma\left[\mathcal{C}\right]\mathcal{Q}\mathcal{Q}^\top\sigma\left[\mathcal{C}\right]^\top ,
\end{align*}
where $\mathcal{K}$ is the kernel network (nonlinear function) in GKN operating on edges, while $\sigma$ is the nonlinear functions in VCNN operating on the nodes. It is evident that the nonlinearity is applied to edges in GKN and to nodes on VCNN.
The comparison is summarized in Table~\ref{tab:gkn-vcnn-compare}.

\begin{table}[!htb]
\caption{Approximate correspondence between graphical kernel network (GKN) and vector-cloud neural network (VCNN).}
\label{tab:gkn-vcnn-compare}
\centering
\begin{tabular}{|| m{4.2cm} | m{5.1cm} | m{5.1cm} ||}
\hline
\qquad \quad Element & \qquad \qquad \quad GKN & \qquad \qquad \quad VCNN \\ \ChangeRT{1.1pt} 
Ensure rotational invariance & Edge construction$^a$: ${\bm{e}}_{ij}$ in Eq.~\eqref{eq:ri_input} & Pairwise inner product$^b$: $\mathcal{Q}\mathcal{Q}^{\top}$ \\ \hline
Ensure permutational invariance & Plain node averaging: \(\frac{1}{n}\sum_i \bm{z}_i^L\) & Weighted node averaging  \(\mathcal{L} = \frac{1}{n} \mathcal{G}^\top\mathcal{Q}\) \\ \hline
Embody inter-node interactions & Kernel network (nonlinear) with edge input \(\mathcal{K}({\bm{e}}_{ij})\) & Pairwise (linear) projection \(\mathcal{Q}\mathcal{Q}^\top\)\\ \hline
Nonlinearity in network & Graphical convolution with edge-dependent (nonlinear) weights & Node-wise embedding network (Fig.4a), fitting network \(\mathcal{D} \mapsto \tau\)\\ \hline
Degenerated to one layer &
\quad \qquad $\frac{1}{n^2}\mathcal{K}\left[\mathcal{Q}\mathcal{Q}^\top\right]\mathcal{C} + \frac{\mathcal{C}}{n}$
& \quad \qquad $\frac{1}{n^2}\sigma\left[\mathcal{C}\right]\mathcal{Q}\mathcal{Q}^\top\sigma\left[\mathcal{C}\right]^\top$ \\ \hline
\end{tabular}
\footnoterule
\raggedright
{\footnotesize $^a$Not a built-in component of GKN, but can be implemented by constructing rotational invariant edge features}\\
{\footnotesize $^b$Equivalently written as $\mathbf{q}_i^\top \mathbf{q}_j$}
\end{table}

\section{Results}
\label{sec:results}
In this section, we demonstrate and analyze the frame invariance property of both neural operators discussed earlier. Unless otherwise stated, the default stencil size $n$ used to report results is 150, which is close to those used in the original works of GKN and VCNN. The results are presented as contours of the scalar field ($\tau$) as well as in terms of prediction error percentage for the testing datasets at different angles of attack with a transformed frame of reference. The prediction error percentage is defined as the normalized discrepancy between predicted scalar field $\hat{\tau}$ and the corresponding ground truth $\tau$: 
\begin{equation}
    \text { Error}= \frac{\sqrt{\sum_{i=1}^{N}\left|\hat{\tau}_{i}-\tau_{i}\right|^{2}}}{\sqrt{\sum_{i=1}^{N}\left|\tau_{i}\right|^{2}}},
\end{equation}
where $N$ represents the total number of points at which the scalar variable has to be predicted. In the first part of this section, we analyze the predictive performance and frame invariance property of the originally proposed GKN~\cite{anandkumar2020neural}. We demonstrate the effect of using rotational invariant input features for GKN. In the second part of this section, we compare two neural 
operators, VCNN that has frame invariance embedded in the neural operator and GKN for which frame invariance is achieved by introducing rotational invariant input features. Training cost and predictive performance are also analyzed for both neural operators.

\subsection{Frame invariance for GKN}
The originally proposed GKN~\cite{anandkumar2020neural} holds permutational and translational invariance properties, which are inherent properties of graph neural networks. However, such a network does not ensure rotational invariance, which can be critical for scientific problems. Reference systems can be differently aligned for different datasets, as shown in Fig.~\ref{fig:train_test_data}. A robust neural operator should have the ability to accurately predict for the dataset with a reference system different from that for the training dataset. For the specified problem setup of predicting concentration field $\tau$ in a flow over an airfoil, the neural operator should be able to accurately predict $\tau$ independent of the reference system of the dataset. In this regard, the frame-invariant characterization of GKN is analyzed here.

To achieve rotational invariance with GKN, we should use rotational invariant input features. The predictive performance of GKN with (Eq.~\eqref{eq:ri_input}) and without (Eq.~\eqref{eq:nonRI_input}) rotational invariant input features is compared in Table~\ref{tab:gkn_results}, which shows the prediction error percentage for the testing dataset with rotated reference system as compared to the training dataset (Fig.~\ref{fig:train_test_data}). It can be observed that the prediction error percentage for the testing dataset is much worse for GKN without rotational invariant input features. For the testing dataset, the reference system is rotated equally to the AOA for the specific flow. In that regard, as the rotational transformation of the reference system increases (with increasing AOA), the prediction error percentage increases from 8.5\% at 5\degree AOA to 27\% at 35\degree AOA. In comparison, GKN with rotational invariant input features predicts with significantly higher accuracy for the testing dataset at all angles of attack. Moreover, it follows a general trend of neural network models, i.e., a better prediction accuracy for interpolating in test conditions than extrapolating in test conditions. 
\begin{table}[!htb]
\centering
\caption{Effect of rotational invariant input features for Graph Neural Network (GKN). Predictive error percentages for two cases correspond to Graph Neural Network (GKN) with and without rotational invariant (RI) input features. Results are computed for stencil size of 150. The training dataset comprises of flow data for three angles of attack (10\degree, 20\degree, 30\degree) \label{tab:gkn_results}}
\begin{tabular}{|| p{2.7cm} | >{\centering\arraybackslash}m{2.5cm} | >{\centering\arraybackslash}m{2.0cm} | >{\centering\arraybackslash}m{2.0cm} | >{\centering\arraybackslash}m{2.0cm} | >{\centering\arraybackslash}m{2.0cm} ||} \hline
 &  & \multicolumn{2}{c|}{
 \begin{tabular}{>{\centering\arraybackslash}m{4.0cm}}
Test Error -- Interpolation \\ \hline
\end{tabular}
 } & \multicolumn{2}{c||}{
\begin{tabular}{>{\centering\arraybackslash}m{4.0cm}}
Test Error -- Extrapolation \\ \hline
\end{tabular}
} \\ 
\; \; NN Model & Training Error & AOA = 15\degree & AOA = 25\degree & AOA = 5\degree & AOA = 35\degree \\ \ChangeRT{1.1pt}
GKN, no RI & 0.27\% & 16.4\% & 23.5\% & 8.5\% & 27.0\% \\ \hline
GKN, with RI & 0.33\% & 1.7\% & 1.4\% & 2.7\% & 2.6\% \\ \hline
\end{tabular}
\end{table}

\begin{figure}[!htb]
    \centering
    \includegraphics[width=0.99\textwidth]{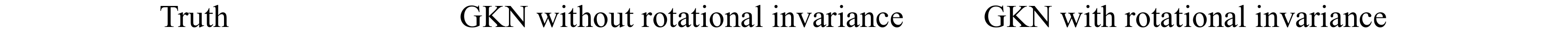}
    \subfloat[AOA=5\degree]{
    \includegraphics[width=0.29\textwidth]{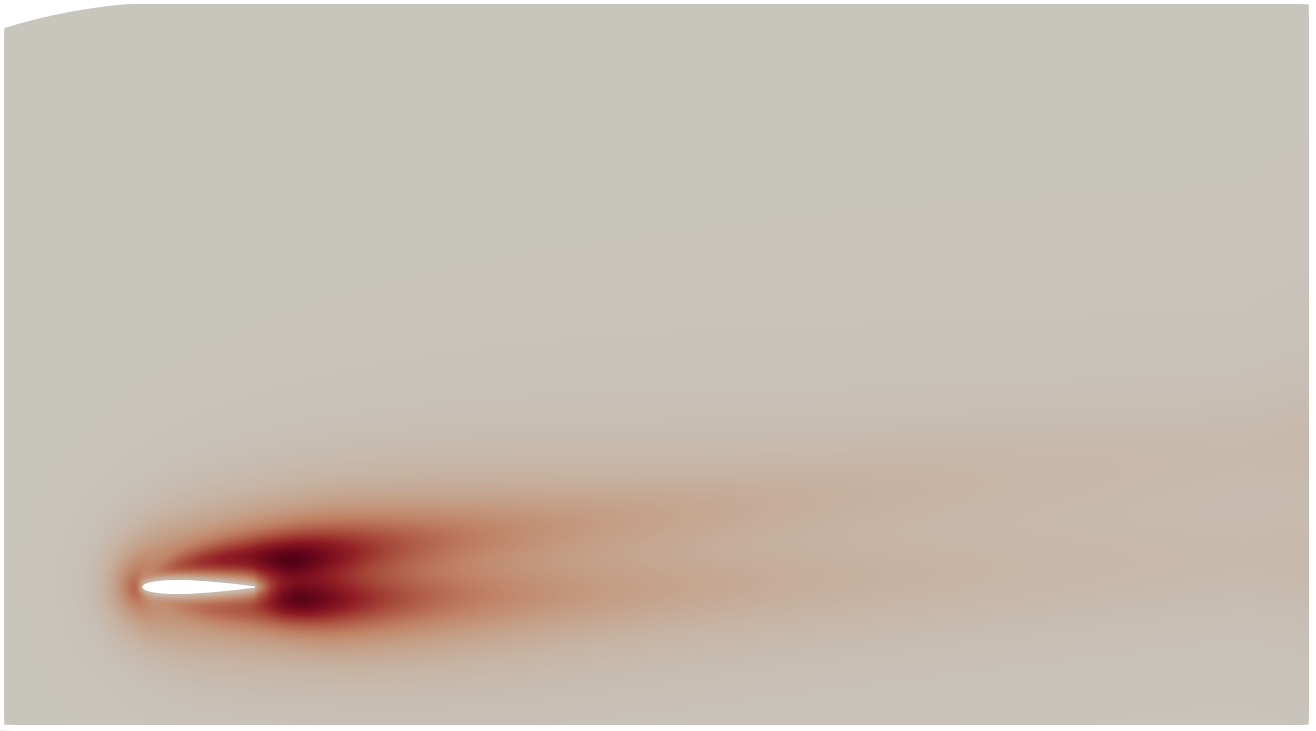}
    \includegraphics[width=0.29\textwidth]{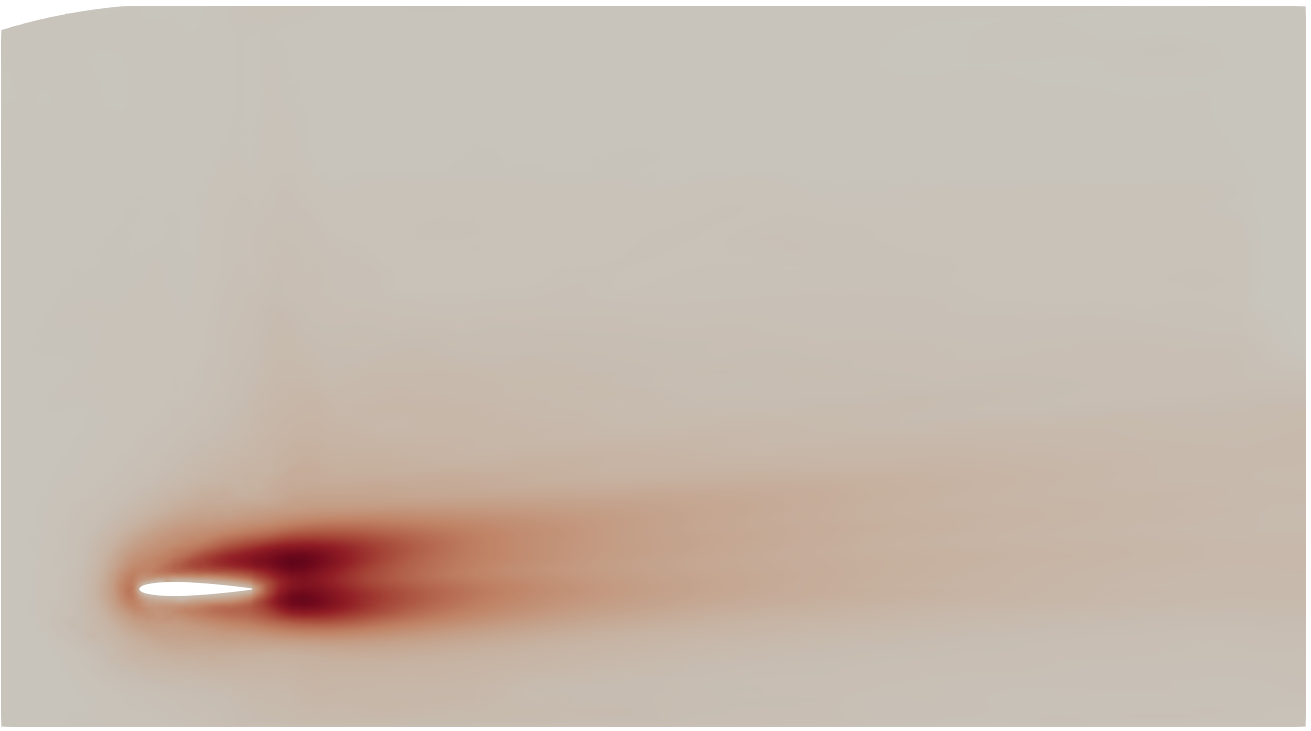}
    \includegraphics[width=0.41\textwidth]{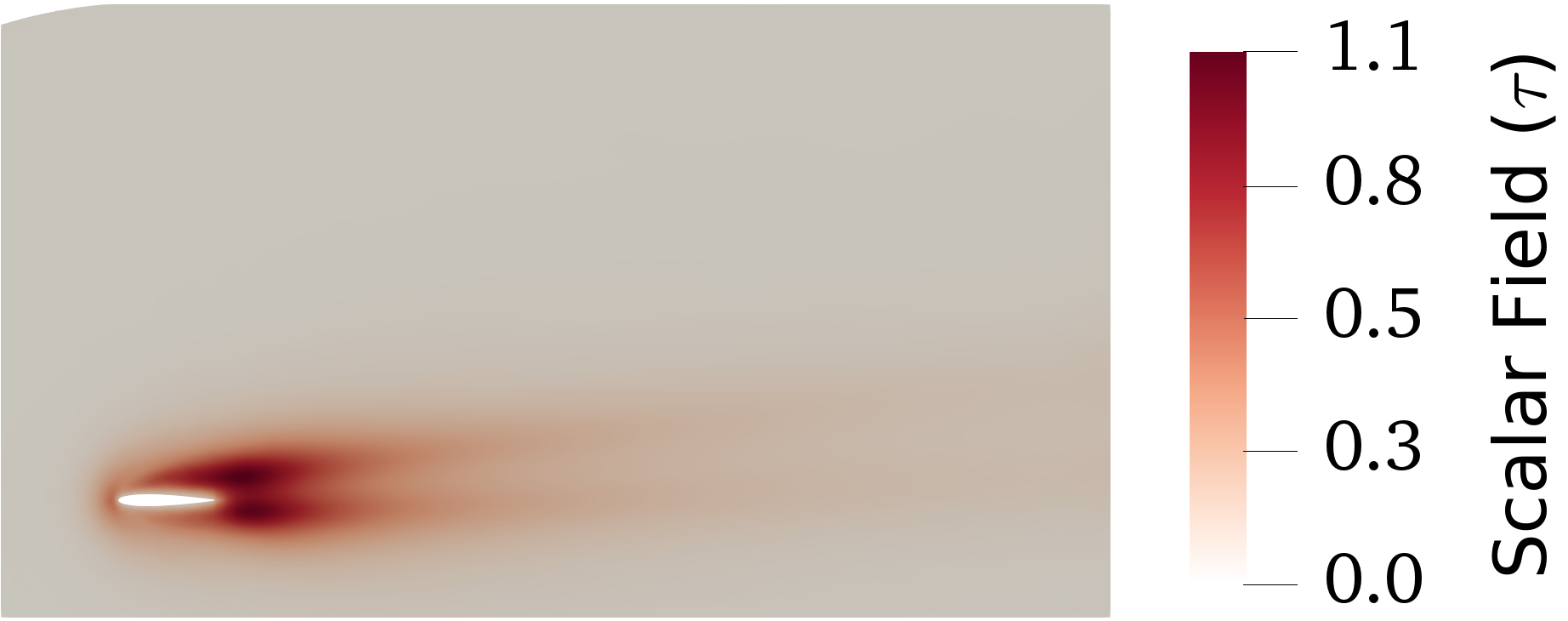}} \\
    \subfloat[AOA=15\degree]{
    \includegraphics[width=0.29\textwidth]{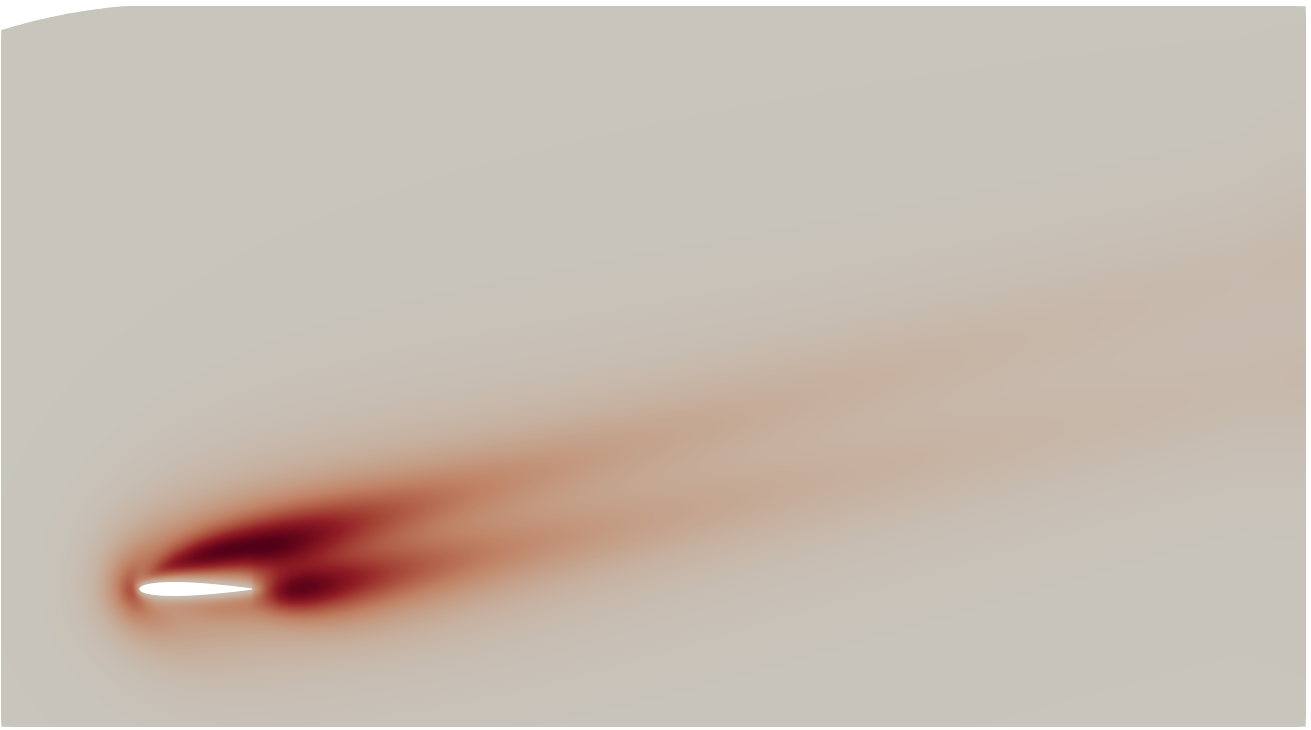}
    \includegraphics[width=0.29\textwidth]{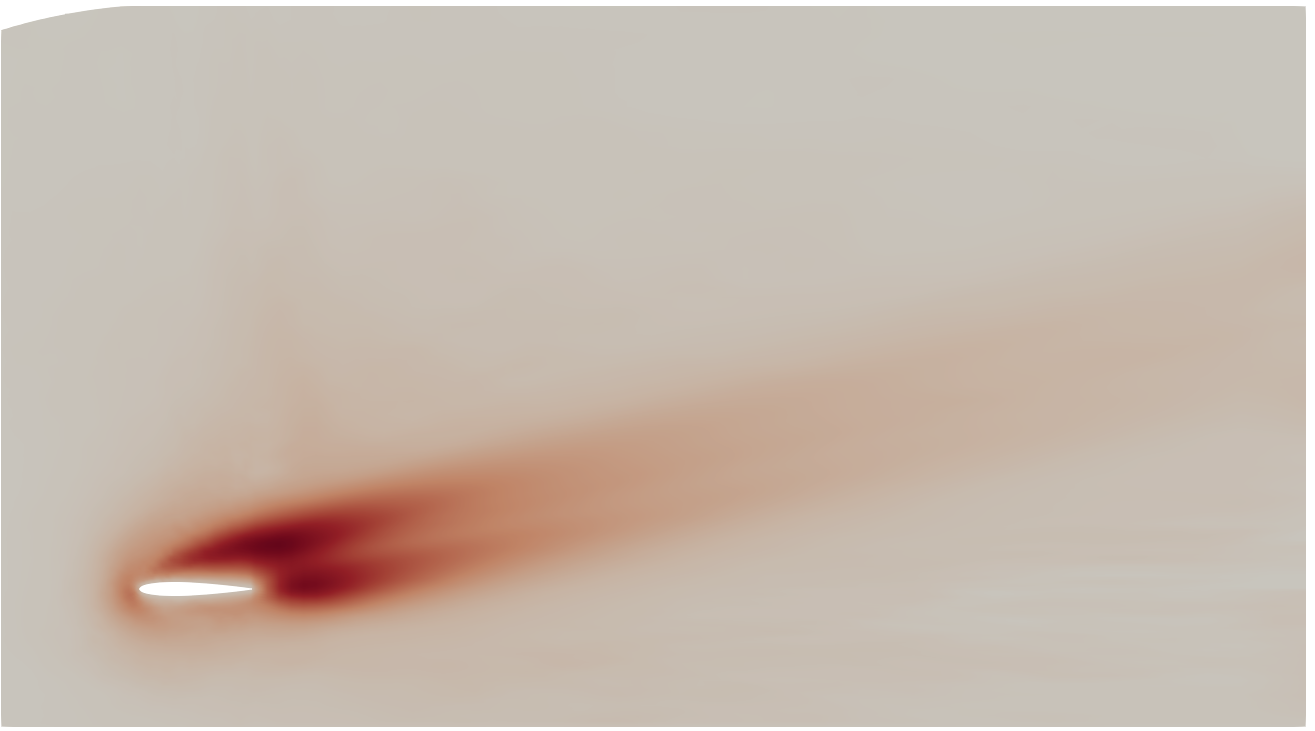}
    \includegraphics[width=0.41\textwidth]{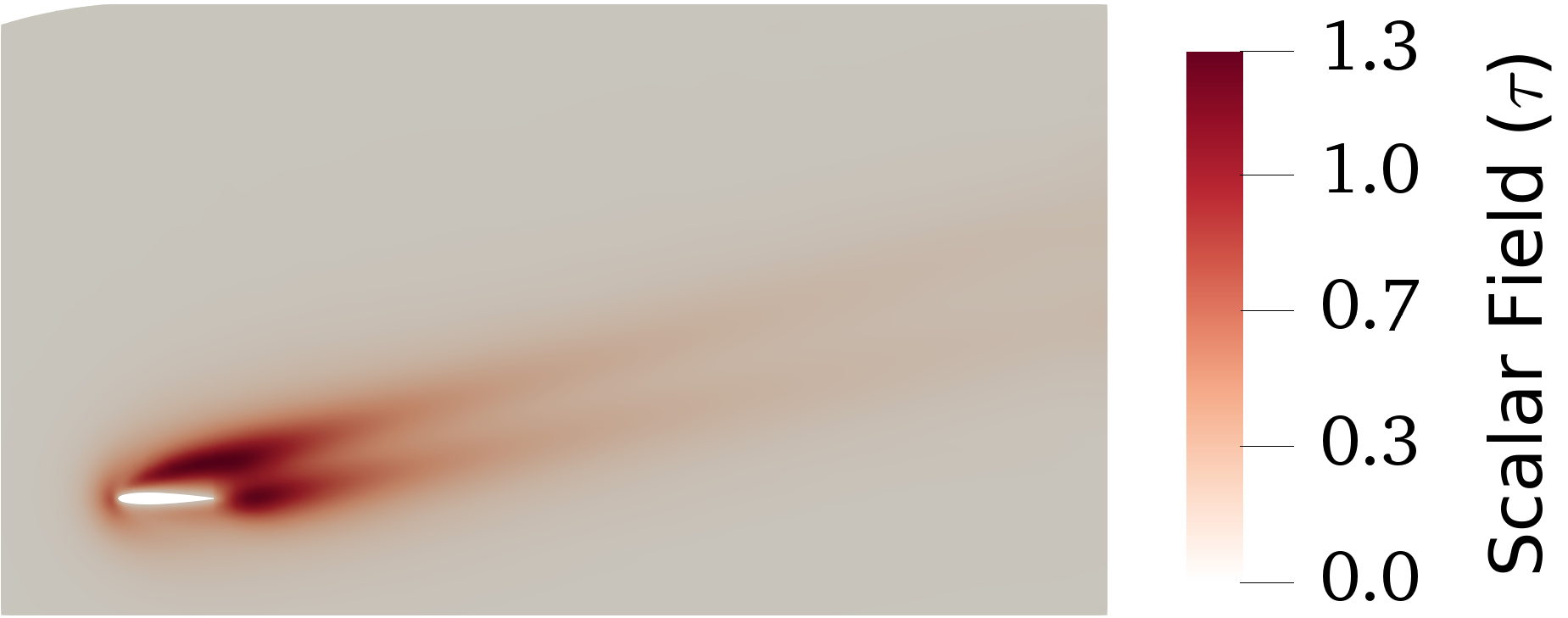}} \\
    \subfloat[AOA=25\degree]{
    \includegraphics[width=0.29\textwidth]{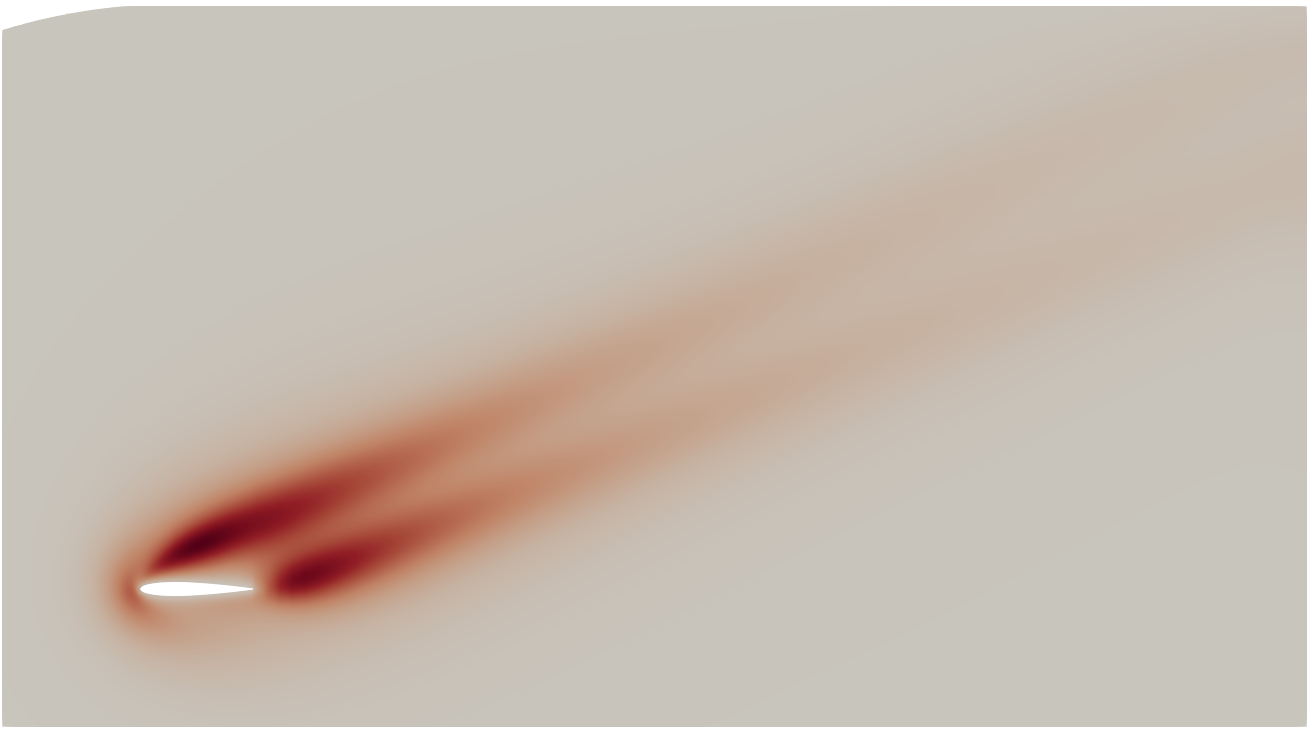}
    \includegraphics[width=0.29\textwidth]{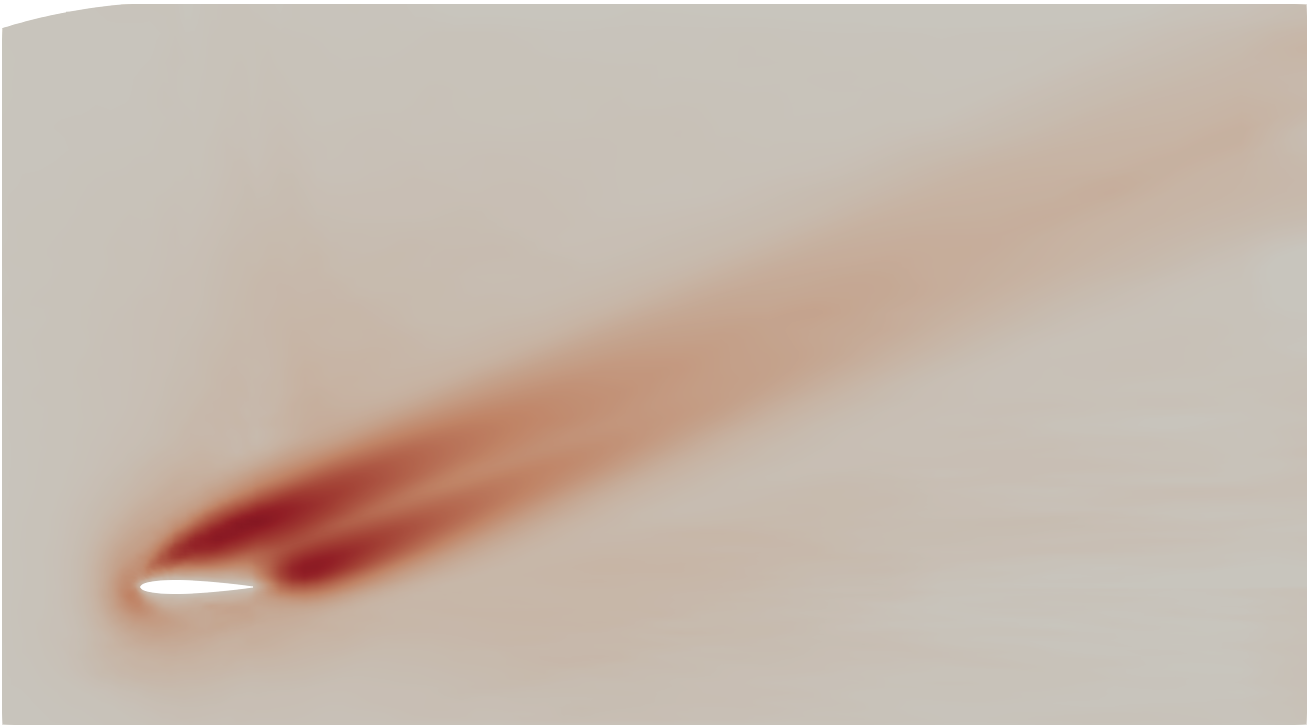}
    \includegraphics[width=0.41\textwidth]{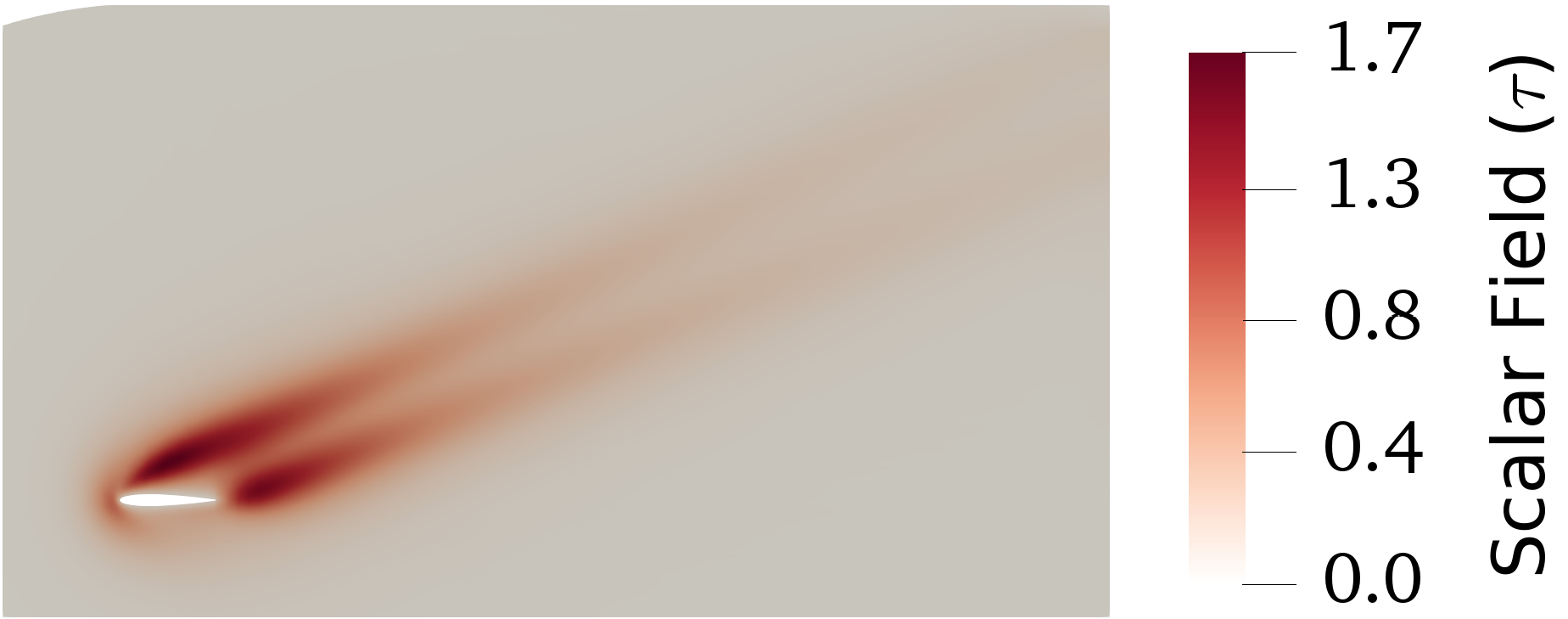}} \\
    \subfloat[AOA=35\degree]{
    \includegraphics[width=0.29\textwidth]{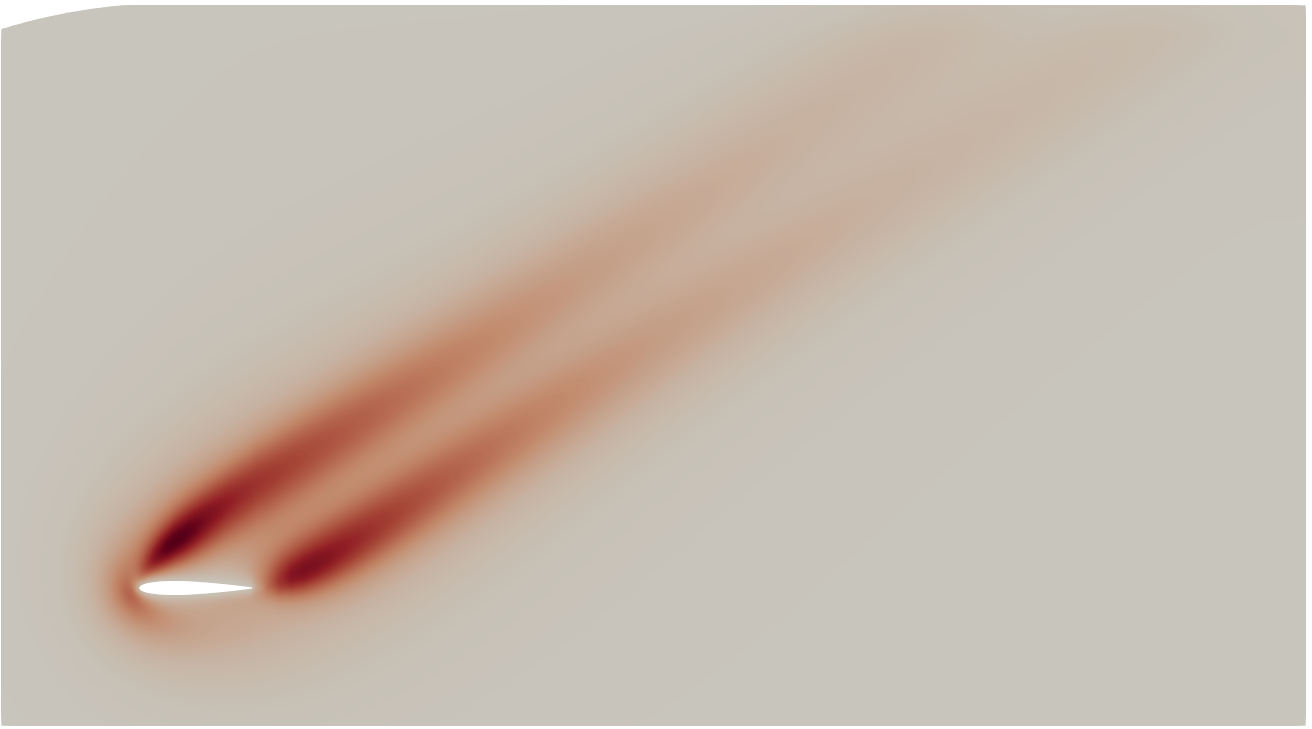}
    \includegraphics[width=0.29\textwidth]{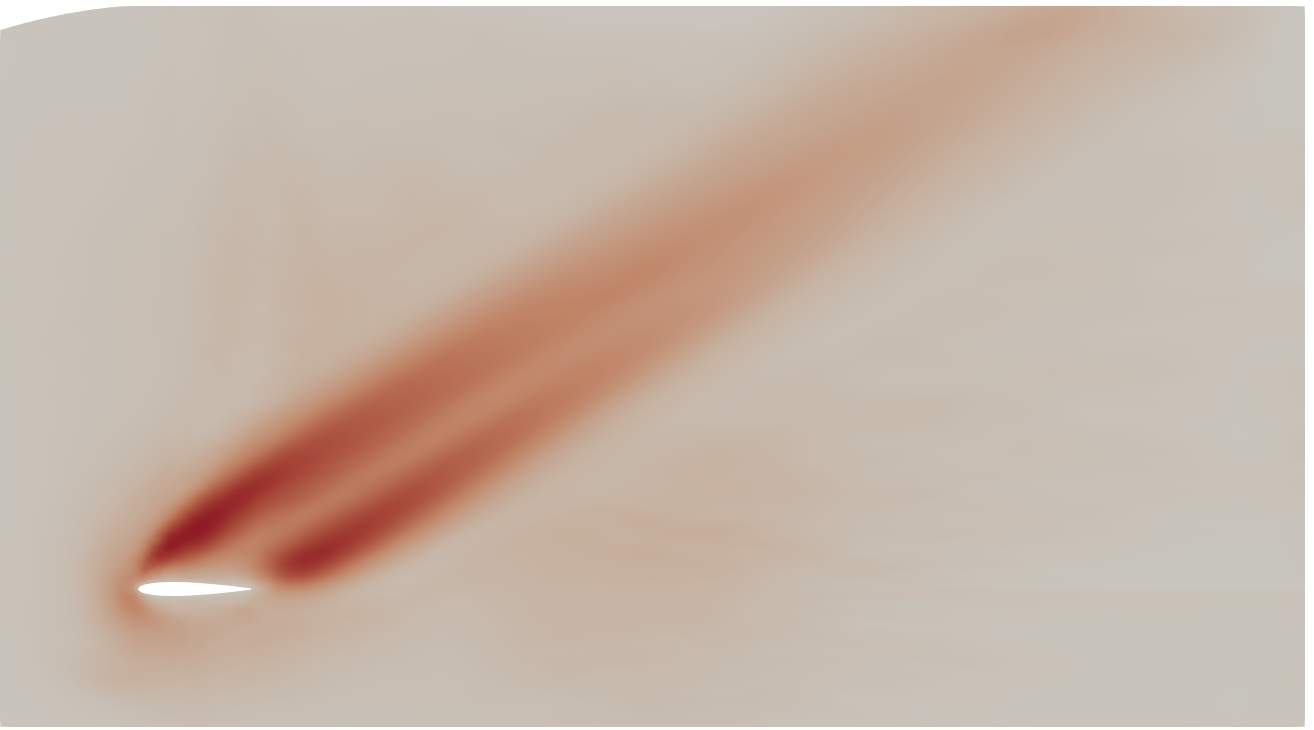}
    \includegraphics[width=0.41\textwidth]{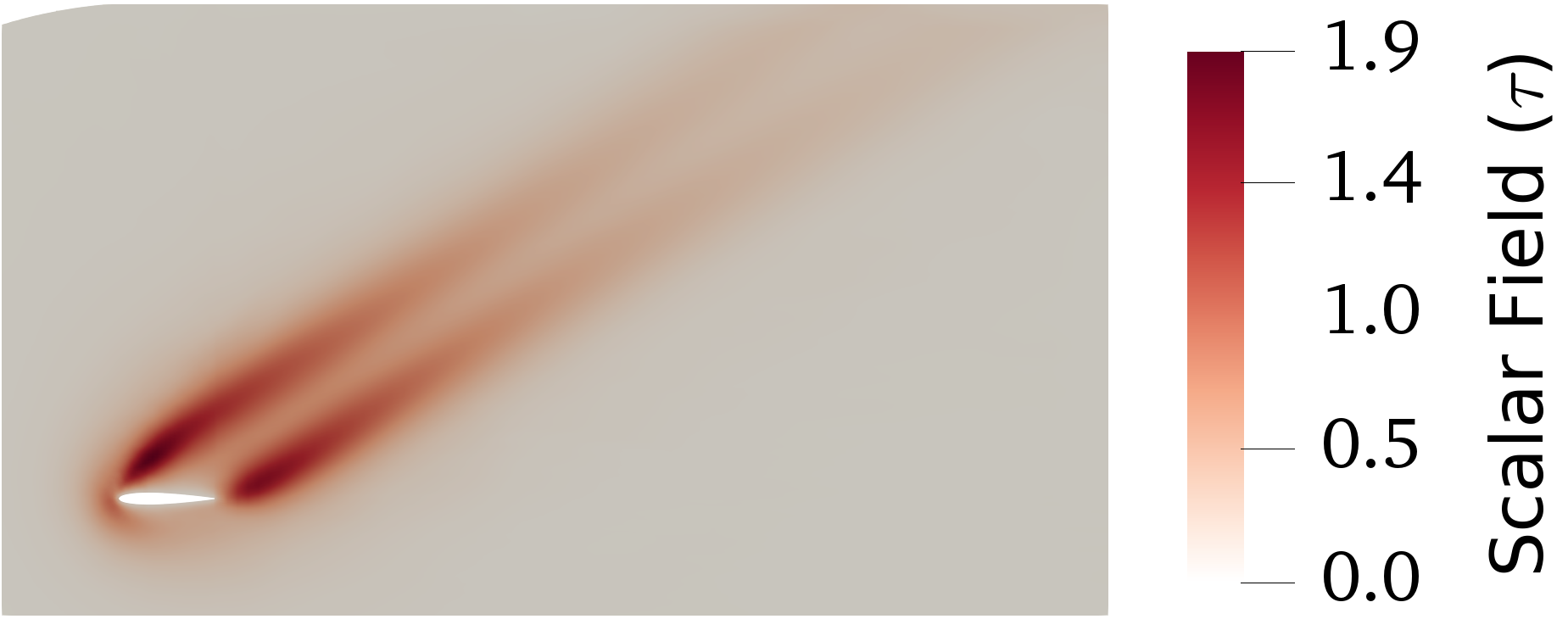}} \\
    \caption{Comparison of graph kernel network predictions with and without rotational invariant input features. Plots show scalar transport field ($\tau$) contours for four testing datasets at different airfoil's angle of attack (AOA).}
    \label{fig:contours_no_RI}
\end{figure}

The contours of the predicted scalar field $\tau$ are shown in Fig.~\ref{fig:contours_no_RI}, where predictions of GKN with and without rotational invariant input features are compared with a true scalar field. In the contours of GKN without rotational invariant input features, erroneously predicted streaks of concentration tracer spread throughout the domain. Moreover for this case, two distinct streams of concentration tracer from upper and lower airfoil surfaces are not clearly distinguishable in the downstream flow. In comparison, the use of rotational invariant input features results in contours that are significantly accurate and imitate those of the true scalar field. 
These results show that the frame invariance can be achieved with the GKN by introducing rotational invariant input features and such a setup provides a promising candidate for frame-invariant neural operators.

\begin{figure}[!htb]
    \centering
    \subfloat[True concentration field ]{\includegraphics[width=0.42\textwidth]{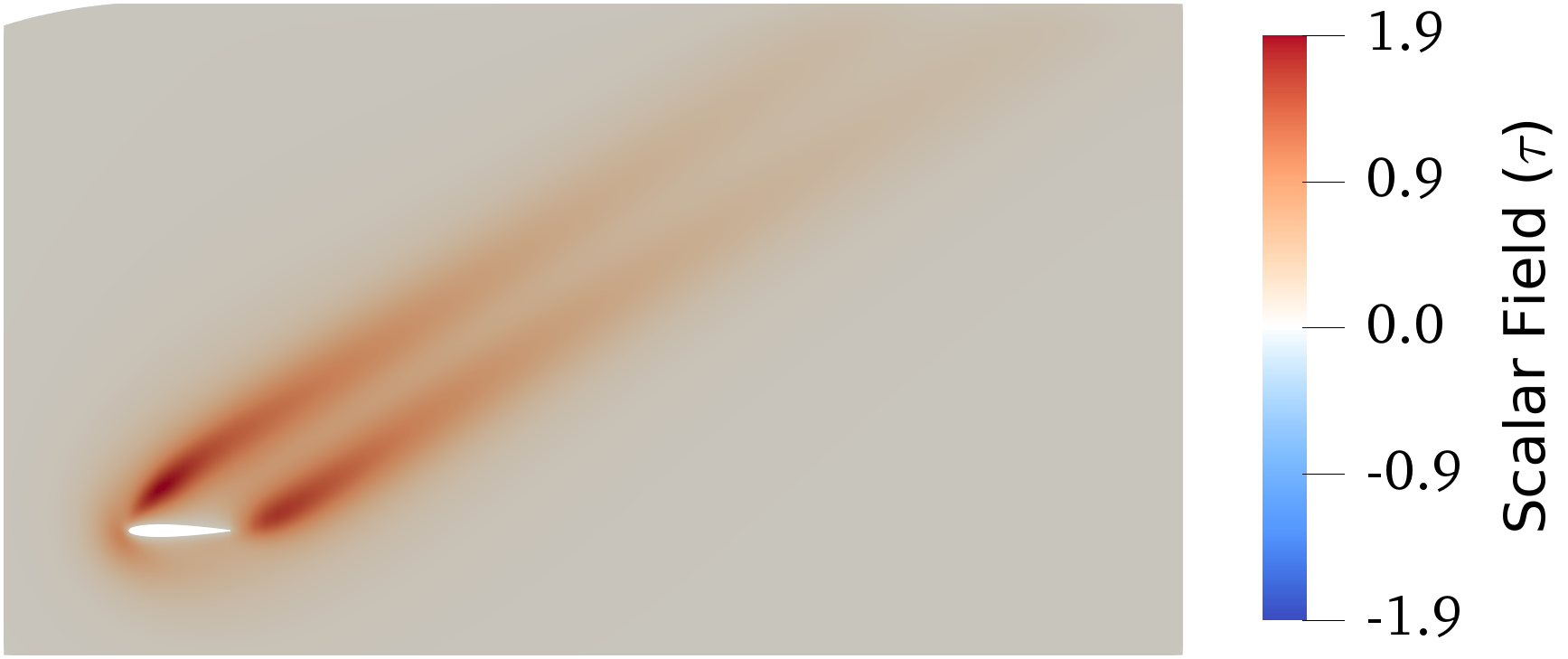}} \qquad \qquad 
    \subfloat[Reference frame rotation = $0\degree$ ]{\includegraphics[width=0.42\textwidth]{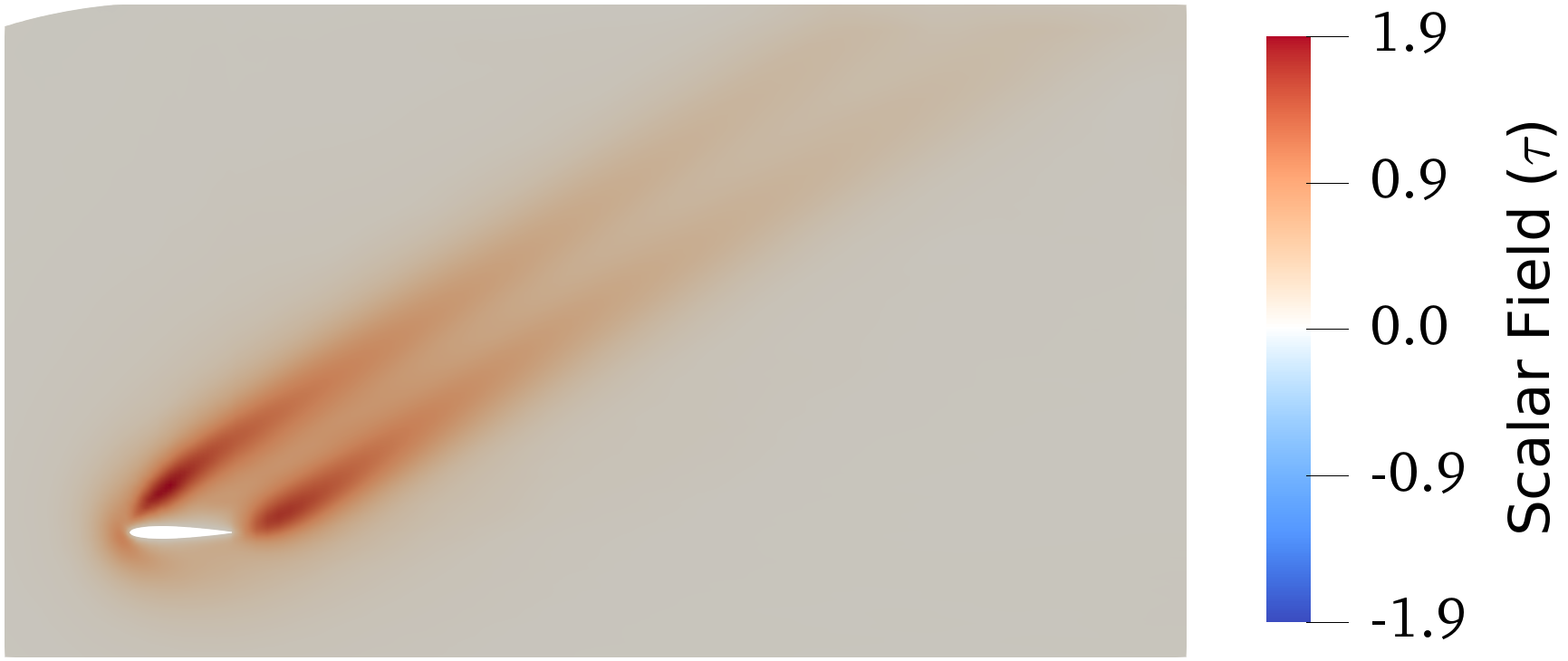}} \\ 
    \subfloat[Reference frame rotation = $35\degree$]{\includegraphics[width=0.42\textwidth]{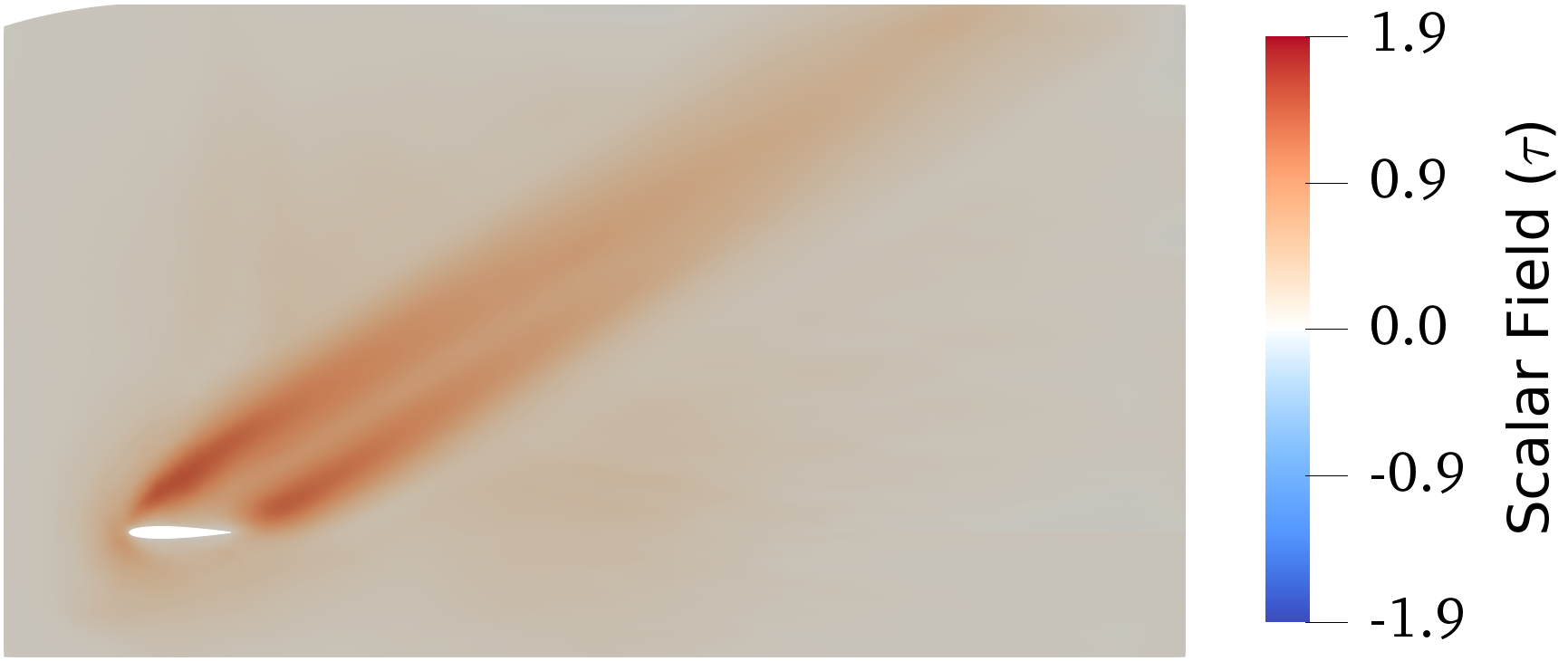}} \qquad  \qquad
    \subfloat[Reference frame rotation = $70\degree$]{\includegraphics[width=0.42\textwidth]{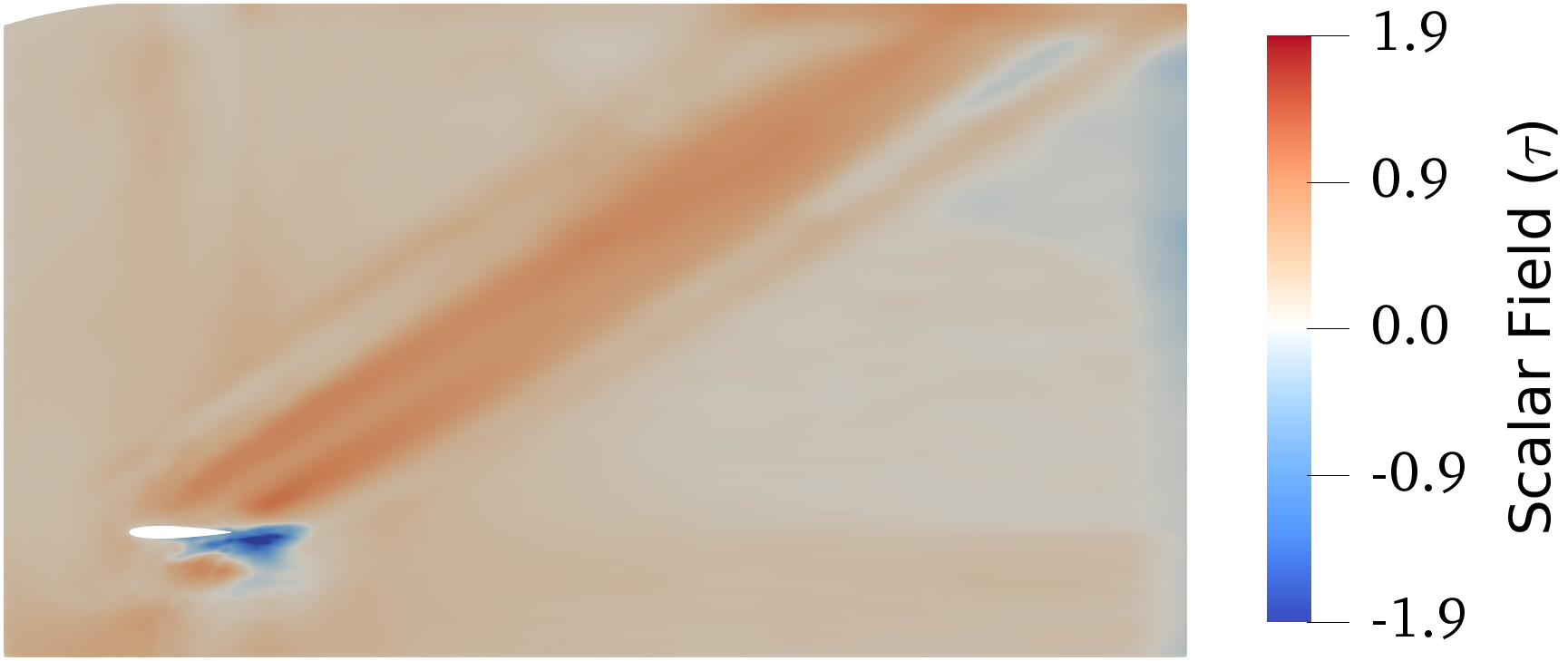}}  \\ 
    \subfloat[Reference frame rotation = $90\degree$]{\includegraphics[width=0.42\textwidth]{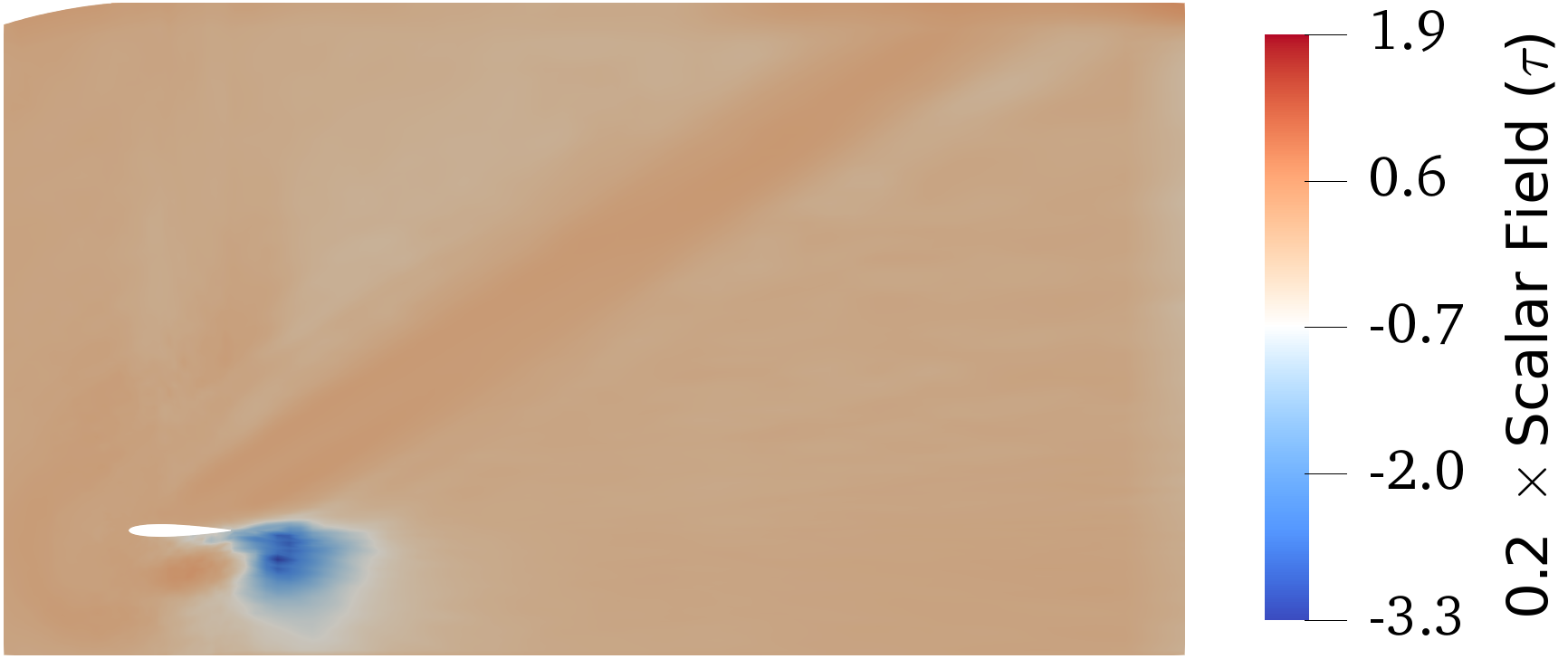}} \qquad  \qquad
    \subfloat[Reference frame rotation = $180\degree$]{\includegraphics[width=0.42\textwidth]{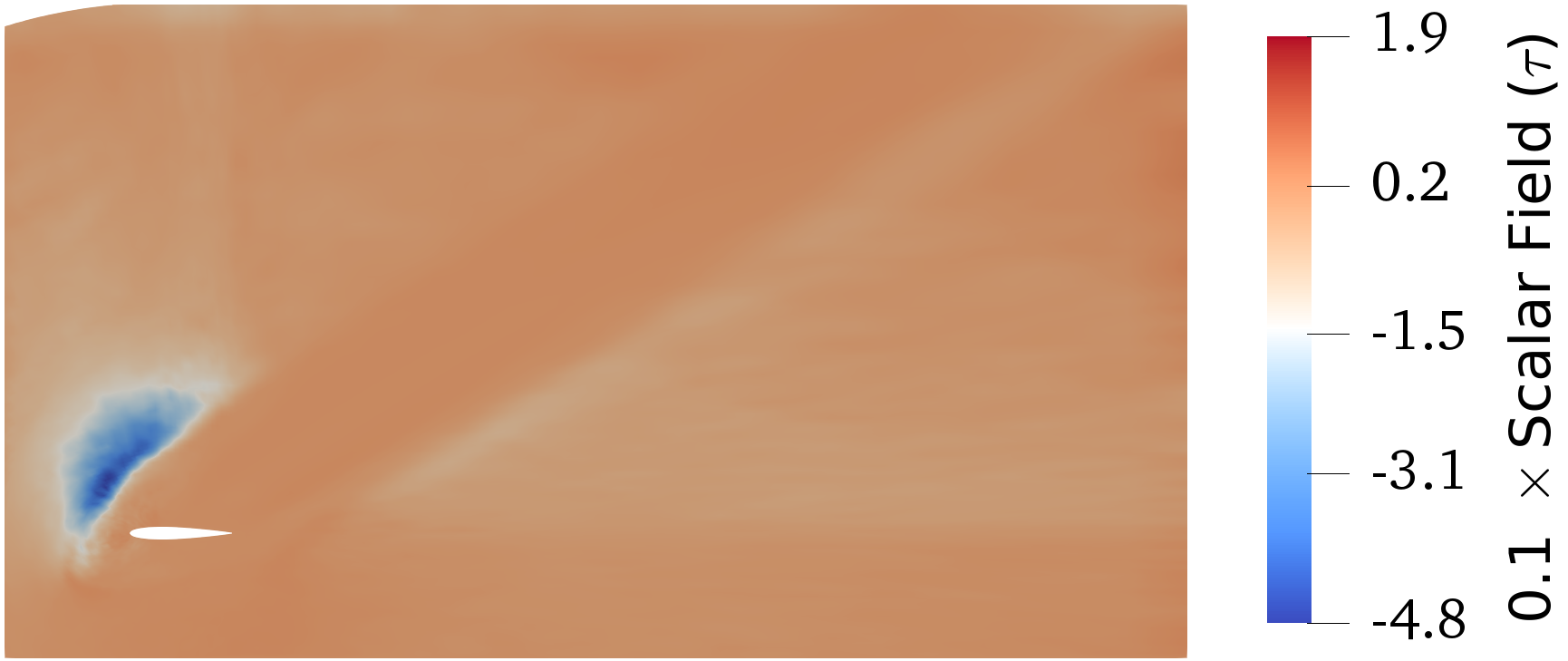}} \\
    \caption{Contours of scalar field $\tau$ predicted by graph kernel network without rotational invariance with input data reference frame rotated by arbitrarily selected angles. Results are shown for the testing data corresponding to AOA=$35\degree$. Scalar quantity $\tau$ has been scaled down by factors of 5 and 10 for reference frame rotation angles of $90\degree$ and $180\degree$, respectively.}
    \label{fig:RI_tests}
\end{figure}

To further highlight the significance of frame invariance for neural operators, GKN without rotational invariance is tested in the case of AOA=$35\degree$, with input data that has been transformed/rotated with arbitrarily selected angles. Fig.~\ref{fig:RI_tests} shows the scalar field $\tau$ contours for these results. Reference frame of input data is rotated to different angles ($0\degree$,$\ 35\degree$,$\ 70\degree$, $\ 90\degree$ or $\ 180\degree$) and the corresponding results are compared with a true concentration field. As the reference frame of the input data is transformed with larger rotation angles, the predicted output by GKN without rotational invariance becomes more erroneous. For input data transformed to large rotation angles ($\ 90\degree$ and $\ 180\degree$), the predicted output (Fig.~\ref{fig:RI_tests}(e) and \ref{fig:RI_tests}(f)) is significantly off-scale compared to true scalar field. In contrast, GKN with rotational invariance predicts accurate concentration field, invariant to the transformations applied to the input data. 

\begin{figure}[!htb]
    \centering
    \includegraphics[width=0.67\textwidth]{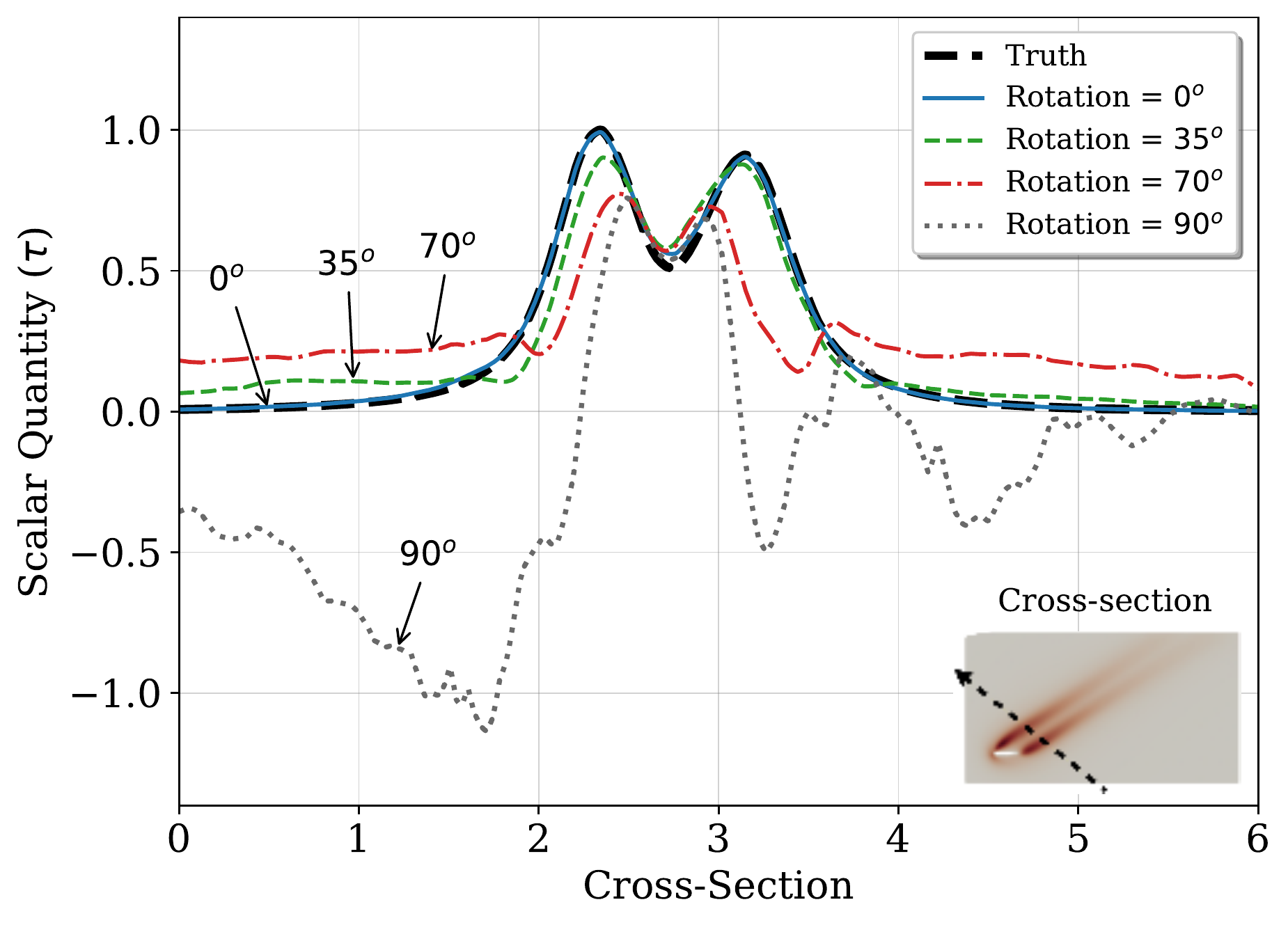}
    \caption{Comparison of profiles along the cross-section in the scalar field $\tau$ predicted by graph kernel network without rotational invariance with input data reference frame rotated by arbitrarily selected angles. The cross-section is shown in the inset, with the arrow indicating the direction along which profiles are plotted. Profiles are normalized with respect to the maximum absolute value of the truth profile. Results correspond to the testing data at AOA=$35\degree$.}
    \label{fig:profiles}
\end{figure}

These results are further analysed by plotting profiles of scalar quantity $\tau$ along a cross-section, as shown in Fig.~\ref{fig:profiles}. The cross-section line has been shown in the inset at the right bottom corner of Fig.~\ref{fig:profiles}, where the arrow indicates the direction along which profiles have been drawn. Profiles plotted along this cross-section show the discrepancies more recognizably. When no rotational transformation ($0^o$) is applied to the input data, the profile agrees well with the truth data profile. As large rotational transformations are applied to the input data, the predicted output profile clearly deviates from the true data, with erroneous negative values for reference frame rotation of $90\degree$. These results signify the importance of the frame-invariance property of the neural operators for scientific problems where differently aligned reference frames are used in different problem cases.

\subsection{Comparison of GKN and VCNN}

With two suitable candidates for frame-invariant neural operators solving scientific problems, we present here a comparison of their predictive performance and computational efficiency. The same training and testing dataset setups are used as in the earlier subsection, i.e., a training dataset at three angles of attack (10\degree, 20\degree and 30\degree) and a testing dataset at four angles of attack (5\degree, 15\degree, 25\degree\ and 35\degree) with transformed coordinate systems equal to the respective flow angles of attack. Table~\ref{tab:gkn_vs_vcnn} shows the comparison of predictive performance for GKN and VCNN. The results of both models are compared using the same stencil size (150), i.e., the same number of sampled data points in the nonlocal region of influence. Furthermore, the number of learnable hyperparameters for GKN and for the fitting network of VCNN are approximately equal ($33000$). Results in Table~\ref{tab:gkn_vs_vcnn} show that the predictive performance of GKN is better than VCNN, despite both learning from the training dataset with comparable accuracy. 
\begin{table}[!htb]
\centering
\caption{Predictive performance comparison of graph kernel network (GKN) and vector cloud neural network (VCNN). Rotational invariant (RI) input features are used for GKN. Results are computed for stencil size of 150. The training dataset comprises of flow data for three angles of attack (10\degree, 20\degree, 30\degree) \label{tab:gkn_vs_vcnn}}
\begin{tabular}{|| p{2.7cm} | >{\centering\arraybackslash}m{2.5cm} | >{\centering\arraybackslash}m{2.0cm} | >{\centering\arraybackslash}m{2.0cm} | >{\centering\arraybackslash}m{2.0cm} | >{\centering\arraybackslash}m{2.0cm} ||} \hline
 &  & \multicolumn{2}{c|}{
 \begin{tabular}{>{\centering\arraybackslash}m{4.0cm}}
Test Error -- Interpolation \\ \hline
\end{tabular}
 } & \multicolumn{2}{c||}{
\begin{tabular}{>{\centering\arraybackslash}m{4.0cm}}
Test Error -- Extrapolation \\ \hline
\end{tabular}
} \\ 
\; \; NN Model & Training Error & AOA = 15\degree & AOA = 25\degree & AOA = 5\degree & AOA = 35\degree \\ \ChangeRT{1.1pt}
GKN - with RI & 0.33\% & 1.7\% & 1.4\% & 2.7\% & 2.6\% \\ \hline
VCNN & 0.31\% & 3.5\% & 3.2\% & 6.8\% & 5.3\% \\ \hline
\end{tabular}
\end{table}

Reasons for this relatively inferior performance of VCNN are not fully comprehensible yet, despite both networks using similar information as input data. The difference in the predictive performance might be attributable to the processing of this information. A simple feed-forward network is used as a fitting network in VCNN for mapping the invariant input information to the scalar quantity $\tau$. In GKN, the graph representation is updated for each node by \emph{message passing} process through kernel function and the resulting node features are updated twice (equal to the depth of the network). With each graph update, features of all the nodes are updated based on the information from their corresponding neighbouring nodes in the graph. The scalar quantity $\tau$ is then predicted by global averaging operation over all the node features. The superior performance of GKN can be attributed to this intricate processing of information. Possible improvement in VCNN requires further investigation and will be a subject of future work. 

\begin{figure}[!htb]
    \centering
    \includegraphics[width=0.99\textwidth]{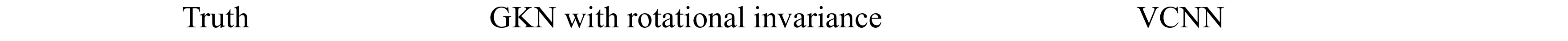}
    \subfloat[AOA=15\degree]{
    \includegraphics[width=0.29\textwidth]{truth_15_150.png}
    \includegraphics[width=0.29\textwidth]{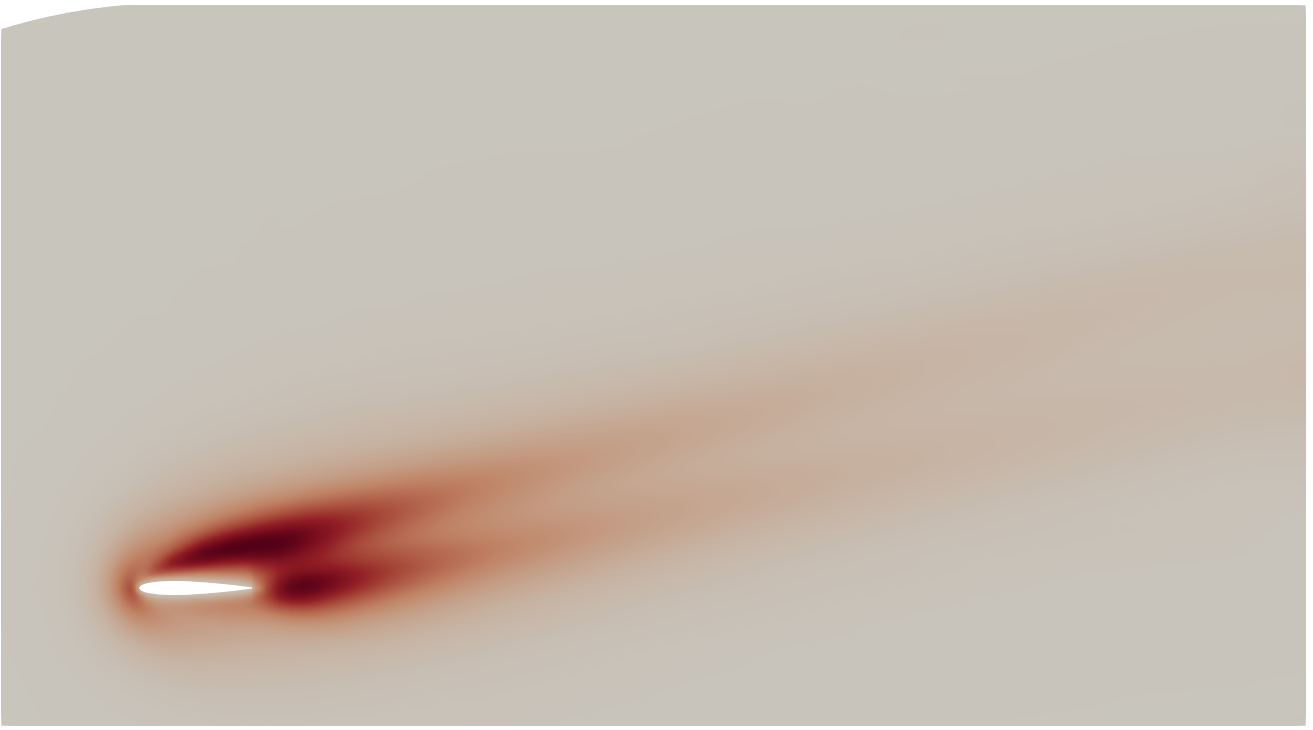}
    \includegraphics[width=0.41\textwidth]{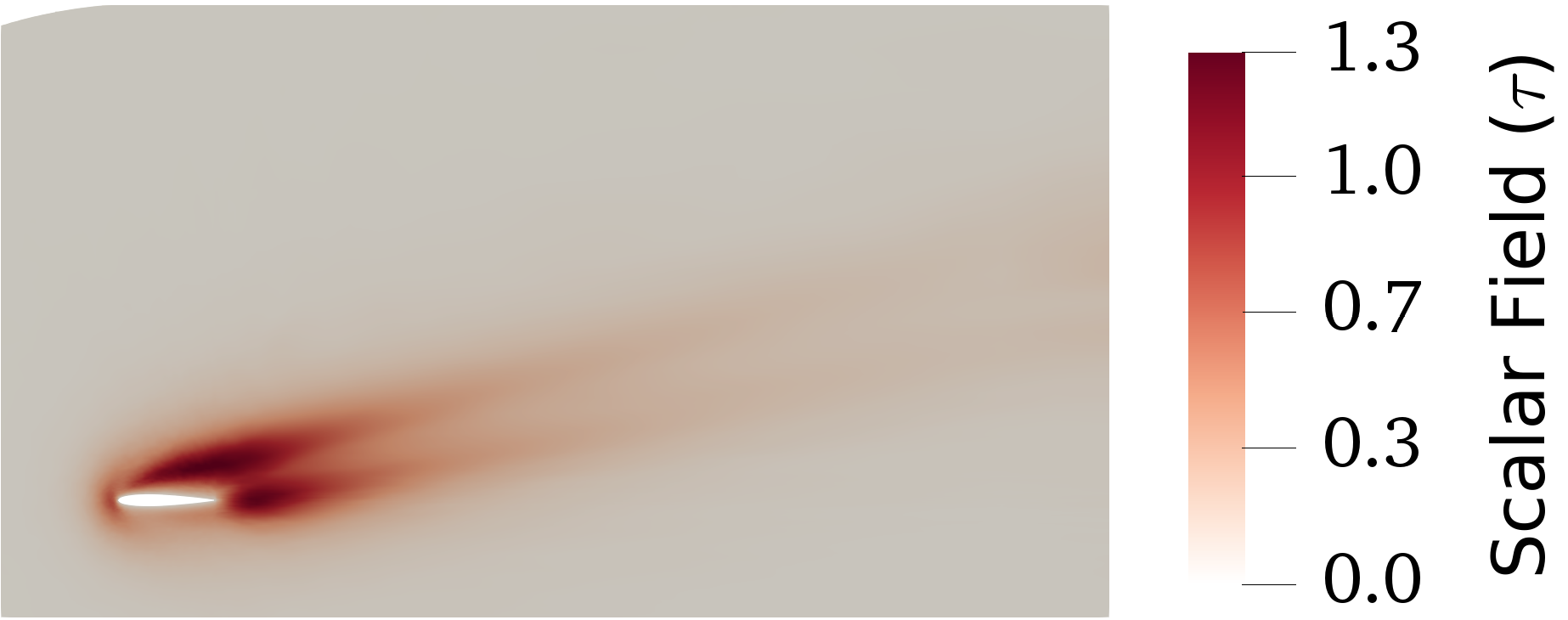}} \\
    \subfloat[AOA=25\degree]{
    \includegraphics[width=0.29\textwidth]{truth_25_150.png}
    \includegraphics[width=0.29\textwidth]{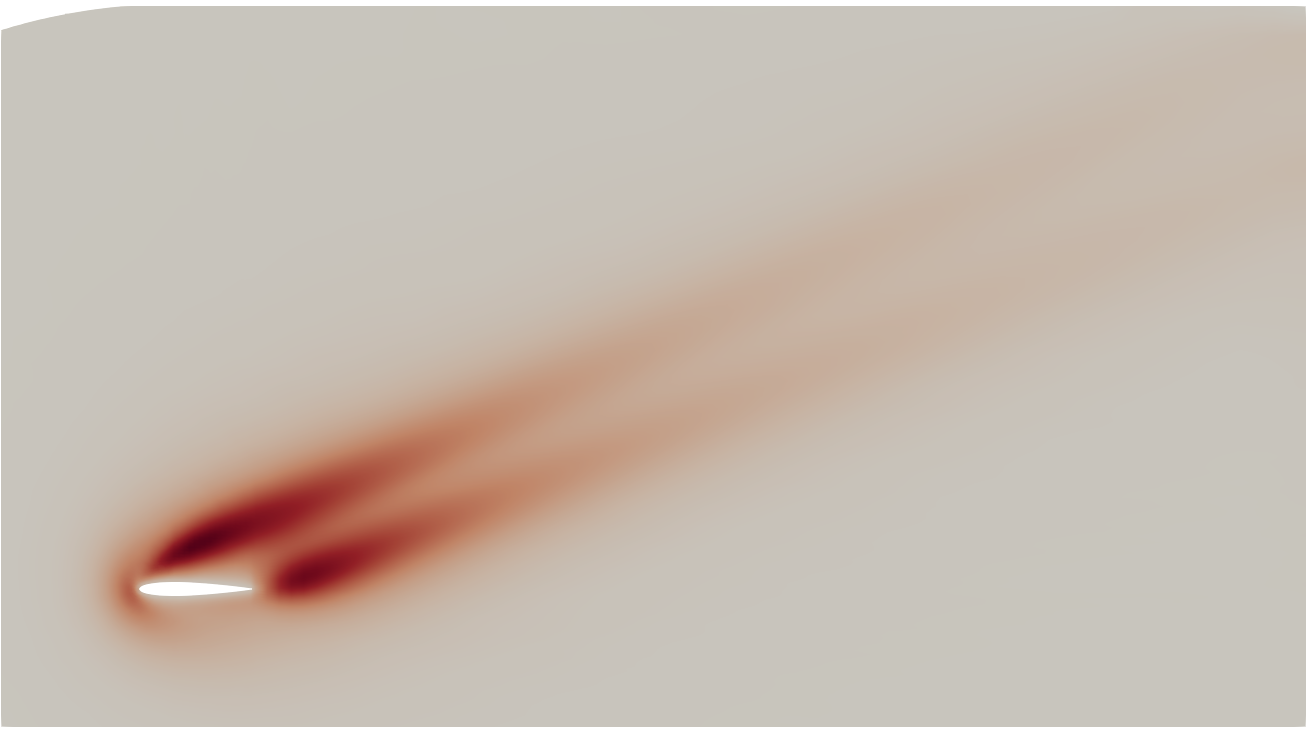}
    \includegraphics[width=0.41\textwidth]{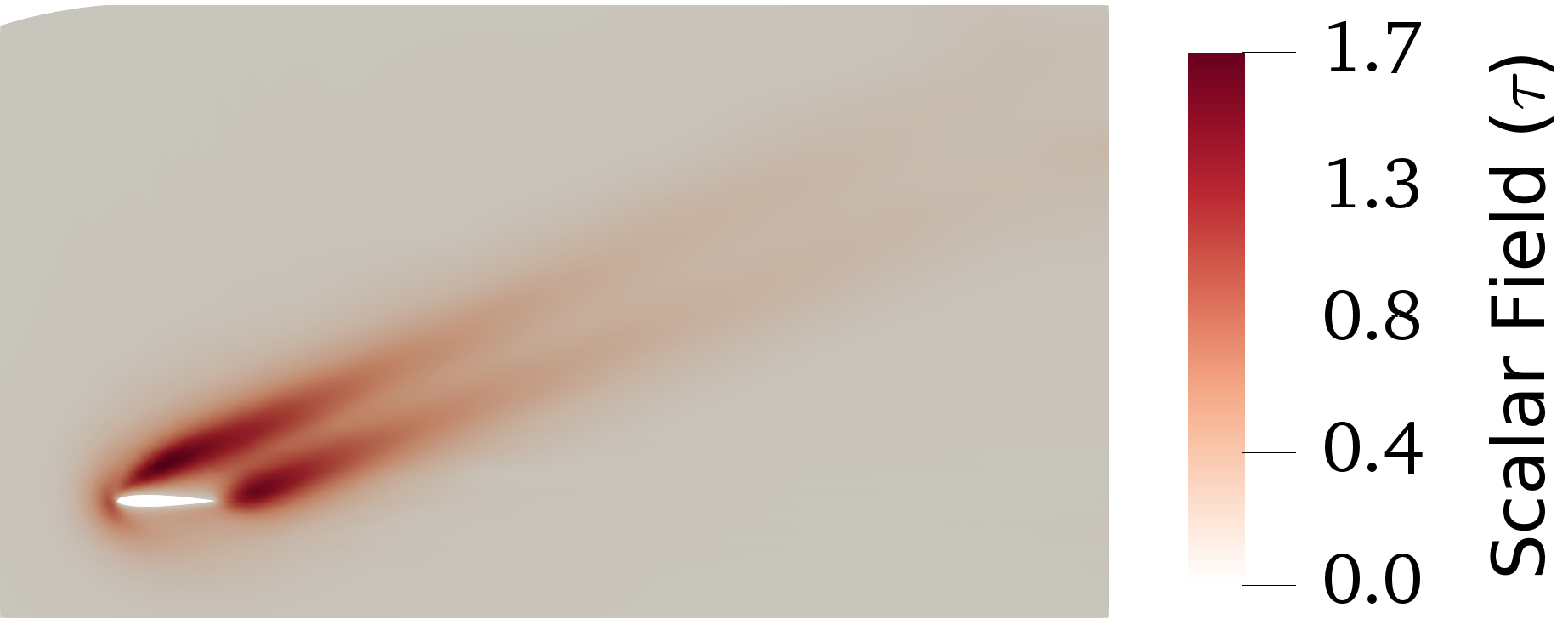}}
    \caption{Scalar transport field ($\tau$) contours for interpolating training data to testing data with respect to airfoil's angle of attack.}
    \label{fig:contours_15_25}
\end{figure}

\begin{figure}[!htb]
    \centering
    \includegraphics[width=0.99\textwidth]{title_gnn_vcnn.pdf}
    \subfloat[AOA=5\degree]{
    \includegraphics[width=0.29\textwidth]{truth_5_150.png}
    \includegraphics[width=0.29\textwidth]{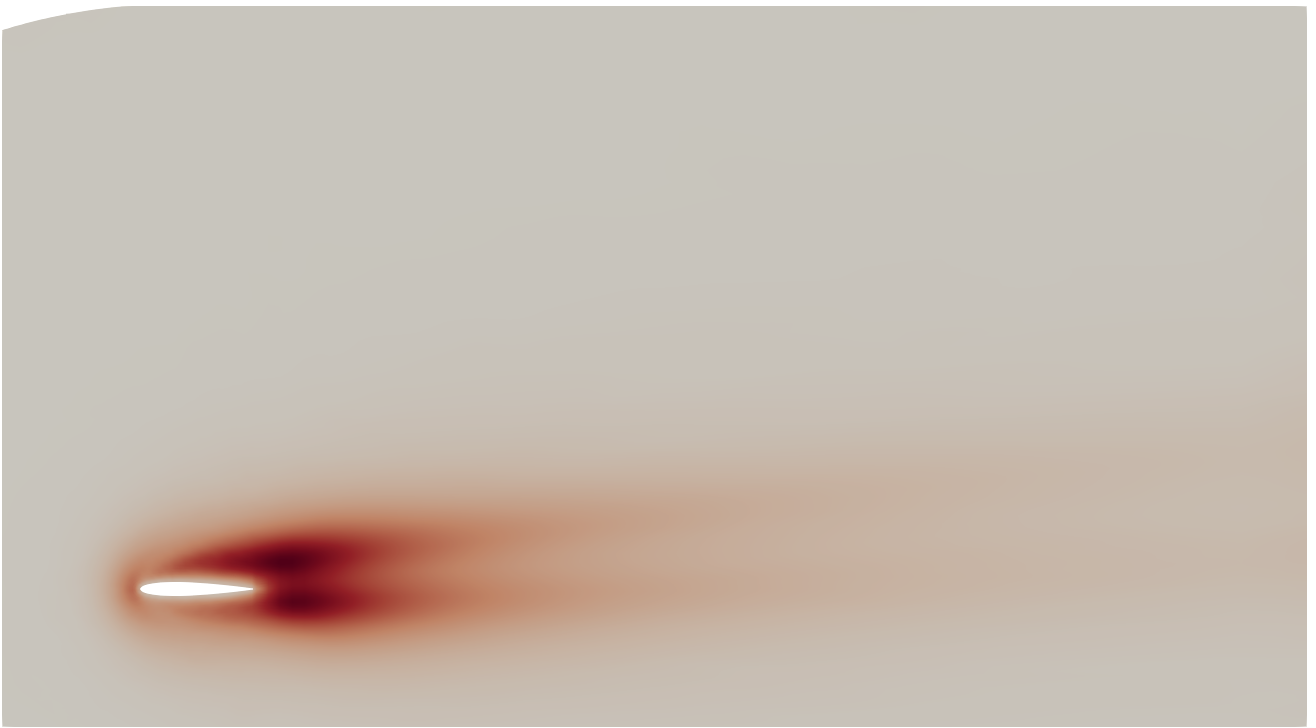}
    \includegraphics[width=0.41\textwidth]{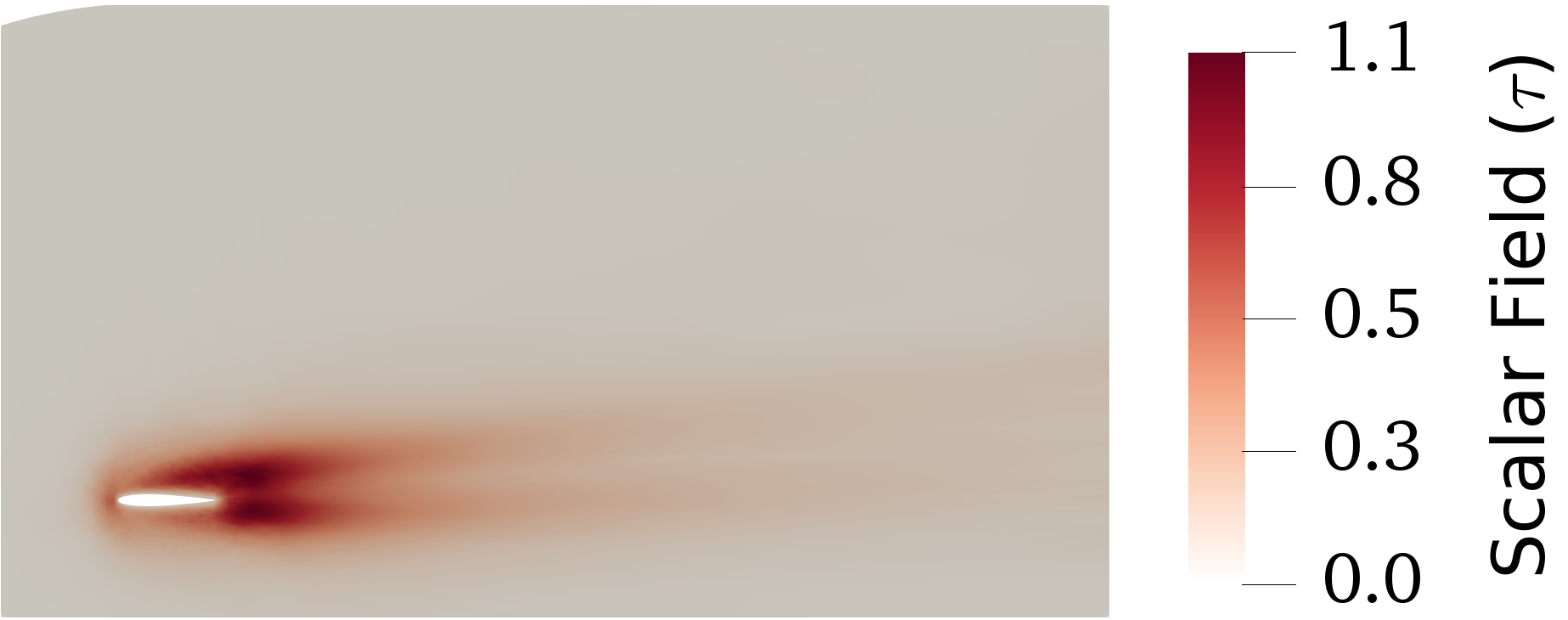}} \\
    \subfloat[AOA=35\degree]{
    \includegraphics[width=0.29\textwidth]{truth_35_150.png}
    \includegraphics[width=0.29\textwidth]{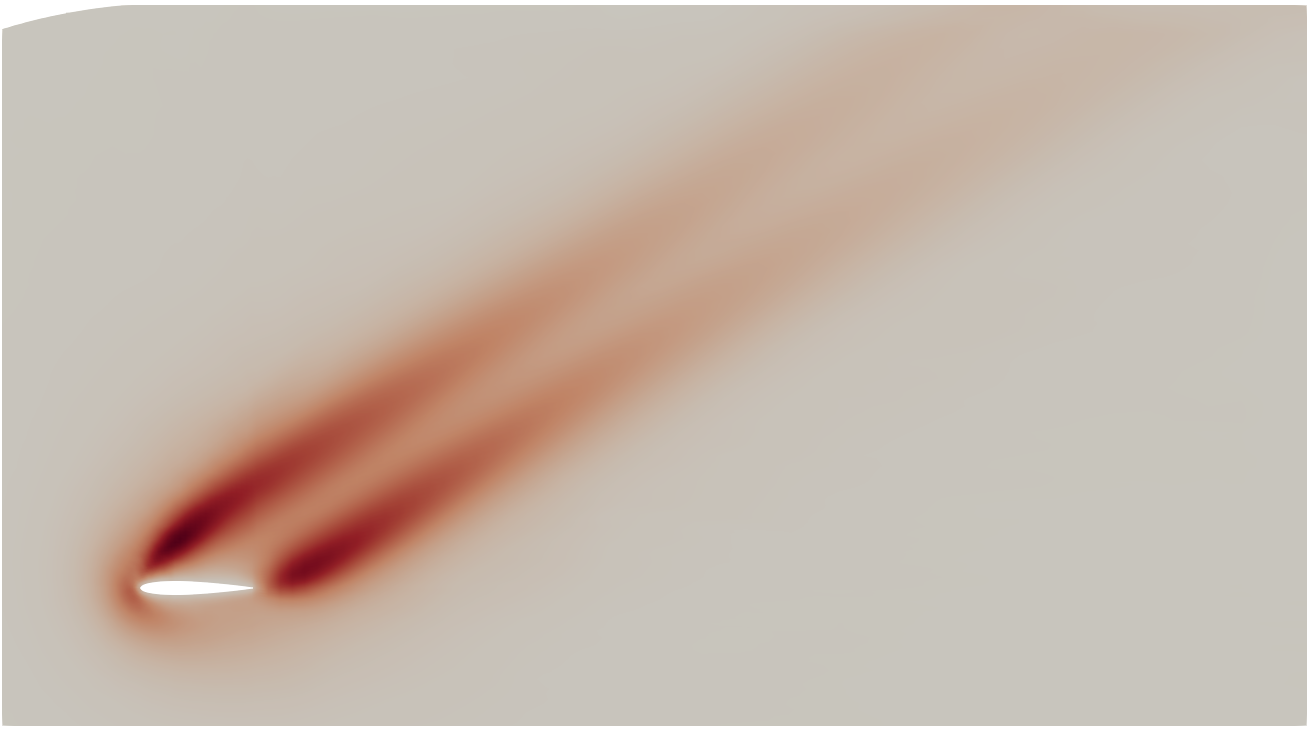}
    \includegraphics[width=0.41\textwidth]{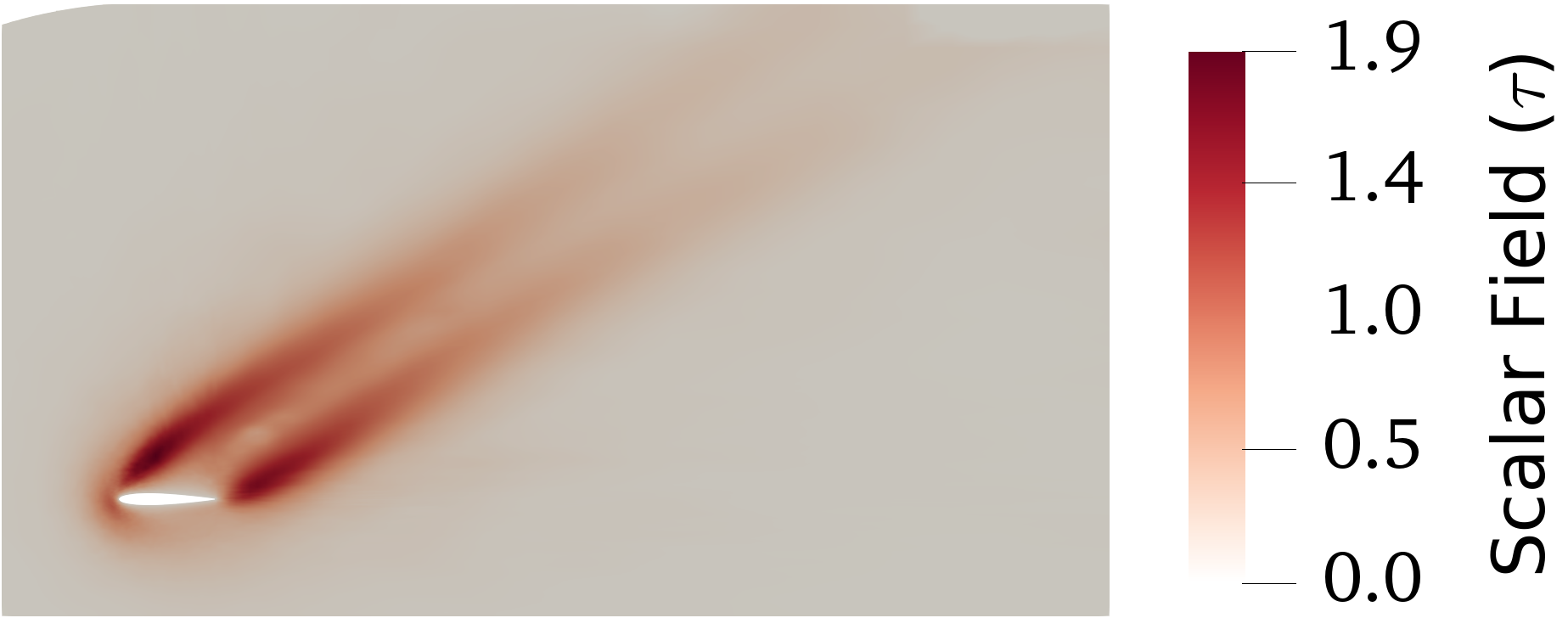}} \\
    \caption{Scalar transport field ($\tau$) contours for extrapolating training data to testing data with respect to airfoil's angle of attack.}
    \label{fig:contours_5_35}
\end{figure}

\begin{figure}[!htb]
    \centering
    \subfloat[Interpolating to test conditions (AOA=15\degree and 25\degree)]{
    \includegraphics[width=0.45\textwidth]{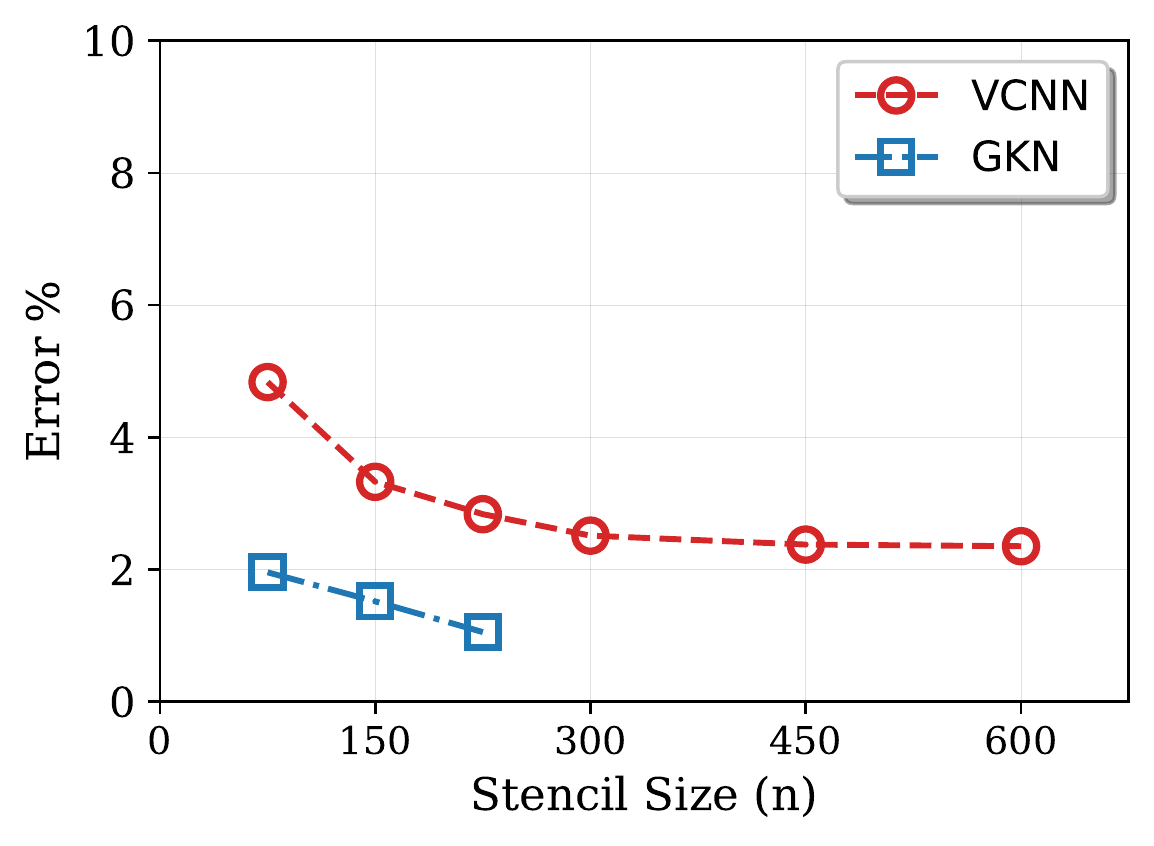}} \qquad
    \subfloat[Extrapolating to test conditions (AOA=5\degree and 35\degree)]{
    \includegraphics[width=0.45\textwidth]{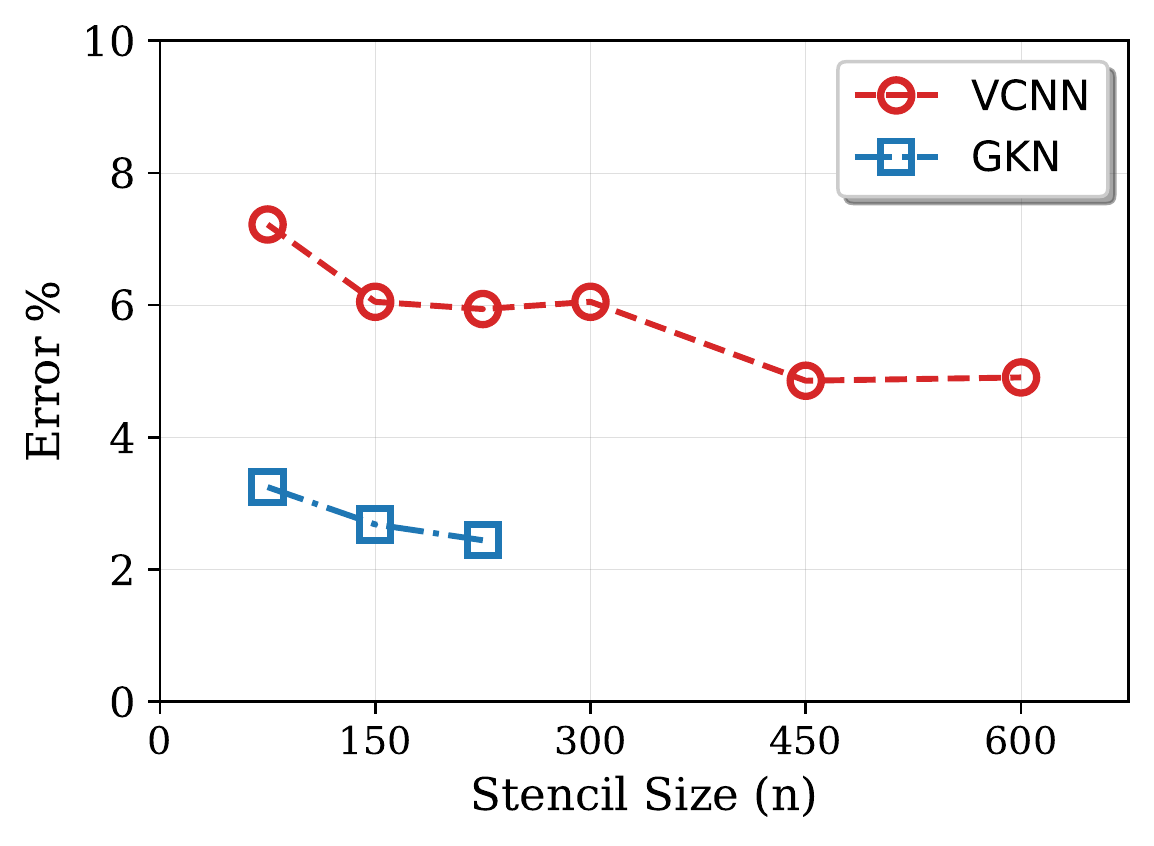}} 
    \caption{Variation of prediction error percentage at testing flow conditions (angles of attack in degrees). GKN for these results is used with rotational invariant input features.}
    \label{fig:err_N}
\end{figure}

Qualitative comparison of the predictions made by both models is shown in Fig.~\ref{fig:contours_15_25} and Fig.~\ref{fig:contours_5_35} for interpolating and extrapolating in  test conditions, respectively. Contours of the scalar field $\tau$ for both models resemble quite well with those of true scalar field $\tau$.
As expected, more information generally leads to better predictive performance. This analysis is shown in Fig.~\ref{fig:err_N} where the predictive error percentage for the testing dataset is plotted as a function of the stencil size. It can be observed that the predictive performance improves with increasing stencil size. For VCNN, the predictive performance shows a converging behavior with increasing stencil size.
For GKN, predictive performance improves with increasing stencil size; however, no observation about the convergence can be made from the given results. 

\begin{figure}
    \centering
    \subfloat[Memory usage for pre-processing the training dataset]{
    \includegraphics[width=0.48\textwidth]{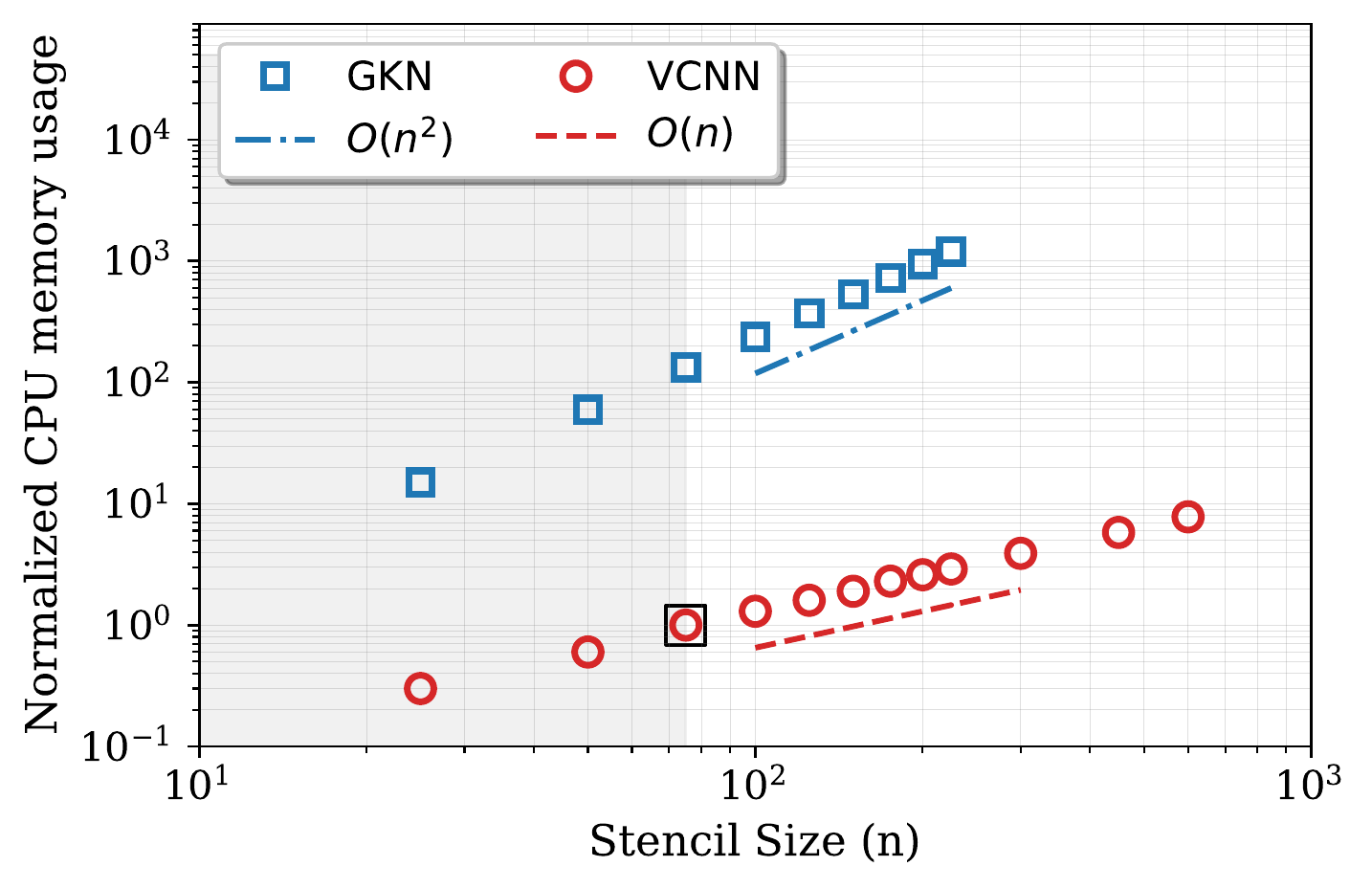}} \quad
    \subfloat[Training time per epoch]{
    \includegraphics[width=0.48\textwidth]{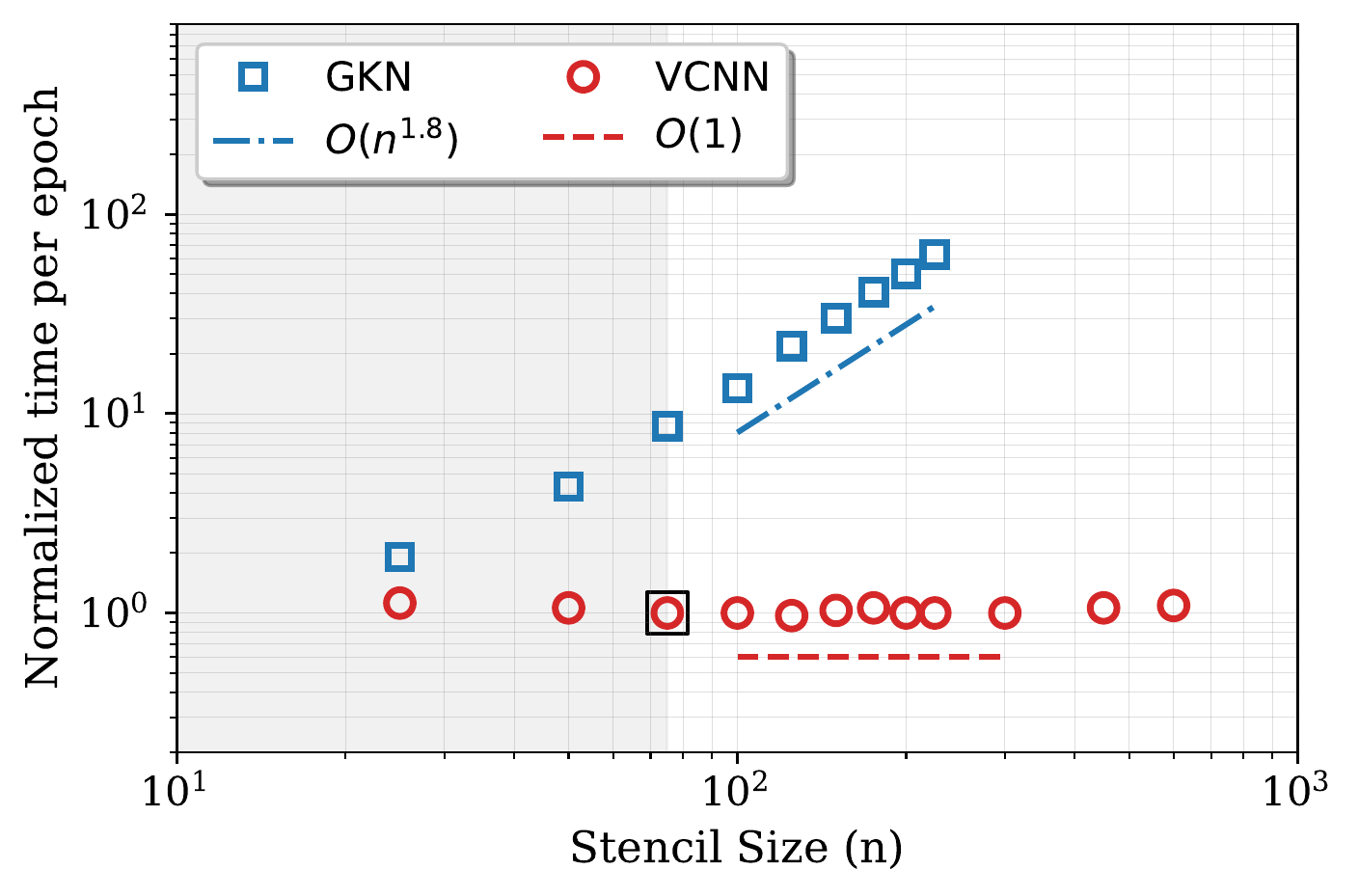}} 
    \caption{Comparison of normalized computational cost for GKN and VCNN. Memory and time costs are normalized with respect to baseline cost of VCNN for $n=75$ (indicated in squares), which is 0.1 gigabytes for the memory usage and 3.3 sec for time per epoch. Gray shaded region indicates small stencil size range which might be unrealistic for intended applications in computational physics. Slope lines illustrate the order of increase in cost with increasing stencil size. The reduction of observed time complexity from memory complexity is mainly due to the parallel computing capacity of GPU.}
    \label{fig:cost}
\end{figure}

In Fig.~\ref{fig:err_N}, although the results for VCNN are shown for the maximum stencil size of the 600,  same are shown for GKN only until 225, because training GKN is prohibitively expensive with larger stencil size. The computational cost in terms of memory usage and training time presents a major disadvantage for graph kernel networks. For the scalar field dataset over an airfoil, a comparison is shown in Fig.~\ref{fig:cost}, where the memory usage and training times are plotted with increasing stencil size. For both neural operators, the time cost is calculated based on the use of one NVIDIA V100 (``Volta'') GPU for training. As the stencil size increases,  the data size for GKN quadratically increases as it involves arranging all the pairwise combinations of the nodes in the graph and the corresponding edge features. For each stencil size $n$, the number of edges has the order of $O(n^2)$. This quadratic increase in data size can also be observed in Fig.~\ref{fig:cost}(a) where CPU memory required to pre-process the training dataset is shown with increasing stencil size. In comparison, the memory usage for VCNN increases linearly. With the same stencil size, memory usage for GKN is approximately two orders of magnitude higher than that of VCNN. For the stencil size of 225, the required CPU memory easily exceeds $100$ GB to pre-process the training dataset for GKN, compared to the requirement of $0.3$ GB for VCNN.
Such an increase in data size is reflected in the training time cost of GKN as well. For VCNN, the linear increase in data size is offset by the parallel computing of GPU. However, the quadratic increase in data size for GKN exceeds the parallel computing capacity of the GPUs, leading to roughly similar order ($\sim O(n^{1.8})$) of increase in training time. With increasing stencil size, the memory requirement and time cost become prohibitively expensive for GKN, which presents a significant disadvantage in comparison with VCNN. On the flip side, for a very small stencil size ($n<75$), the training time cost of GKN is comparable to that of VCNN,
albeit still more expensive in memory.
However, such a stencil size range, illustrated by the shaded gray region in Fig.~\ref{fig:cost}, is not a realistic cloud size for many applications in computational physics.

Beyond the CPU memory cost requirement for pre-processing the data, training of GKN using GPUs also presents a challenging requirement for GPU memory. For the current research work, the training dataset was distributed into large number of batches (800) and each batch is moved to GPU for training one by one due to the limitation of GPU memory. With increasing graph size (stencil size), the requirement would be to distribute the training dataset in an even larger number of batches. For many scientific problems, the graph size could be significantly larger than the current problem, which could lead to a situation where GPU memory requirement for even one graph could prove excessive. Such issues of handling large memory costs for graph kernel networks is in itself a topic of research~\cite{chen2018fastgcn,fey2019,lin2020,jia2020gnnroc}. In this regard, VCNN provides significantly efficient neural operators for scientific problems.

We remark that in~\cite{anandkumar2020neural} the authors proposed to randomly sample $m$ times sub-graphs with node size $n$ from the region to build input--output data pairs. In their cases, the authors observe that parameters $n=200$ and $m=4$ give a good balance between prediction accuracy and computation cost. In the present work, we have chosen $n=150$ and $m=1$. The general guideline for the choice of $n$ and $m$ in GKN still remains elusive and requires further investigation. Meanwhile, the idea of sampling smaller sub-graphs can also be employed by VCNN to improve the training efficiency; see the discussion in~\cite{zhou2021vcnne}.

\section{Conclusion}
\label{sec:conclusion}
Frame-invariant neural operators for transport partial differential equations (PDEs) have been analyzed. Transport of a scalar quantity in a flow over an airfoil is considered to analyze the frame invariance properties of both neural operators. First, graphical kernel network (GKN) model is considered which leverages the structure and properties of graphs to embed nonlocal dependencies. We propose to use invariant input features for the GKN model to guarantee frame invariance. Secondly, we have considered vector-cloud neural network (VCNN) which has both nonlocal dependencies and frame invariance embedded in the model. 

GKN with invariant input features has shown to achieve frame invariance, which is demonstrated for the testing datasets with transformed reference frames. VCNN also demonstrated invariance to transformation in the reference frame; however it showed slightly worse predictive performance as compared to GKN, given the same input data and number of learnable parameters. With increasing stencil size, both models showed improved predictive performance. However, as graph-based networks are intrinsically expensive to train, GKN becomes prohibitively expensive for larger stencil size. In comparison, VCNN demonstrated orders of magnitude better computational efficiency for increasing stencil size. This observation is critical for many scientific problems where large mesh sizes and possibly larger number of feature vectors are used. For such cases, using GKN becomes infeasible and VCNN can be used efficiently as a surrogate model for transport PDEs with embedded invariance properties and nonlocal dependencies. 

Despite the satisfactory performance and computational efficiency of VCNN, a future study is required to address its slightly inferior performance to GKN. Since GKN uses relatively intricate processing of the invariant input data, different options can be explored for the fitting network of VCNN to enhance model accuracy.

\section*{Acknowledgments}
H. Xiao is supported by the U.S. Air Force under agreement number FA865019-2-2204. The U.S. Government is authorised to reproduce and distribute reprints for Governmental purposes notwithstanding any copyright notation thereon.
The computational resources used for this project were provided by the Advanced Research Computing (ARC) of Virginia Tech, which is gratefully acknowledged. 

\appendix

\section{Performance of an improved VCNN}
In this appendix we consider a variant of VCNN for potential performance improvement.
Pairwise projection $\mathcal{Q}\mathcal{Q}^{\top}$ among feature vectors is used to determine the invariant feature matrix $\mathcal{D}$ (Eq.~\ref{eq:D}). Input feature matrix $\mathcal{Q}$ can also be written as a stack of three sub-matrices as:
\begin{equation}
\mathcal{Q} = [ \mathcal{X} \quad \mathcal{U} \quad \mathcal{C} ]
\end{equation}
where sub-matrices $\mathcal{X}$, $\mathcal{U}$ and $\mathcal{C}$ comprises of relative spatial coordinates, velocity vectors and scalar quantities of the point cloud, respectively. Pairwise projection can then be represented as:
\begin{equation}
\mathcal{Q}\mathcal{Q}^{\top} =  \mathcal{X}\mathcal{X}^{\top} + \mathcal{U}\mathcal{U}^{\top} + \mathcal{C}\mathcal{C}^{\top}.
\end{equation}
To retain more information in the invariant feature matrix $\mathcal{D}$, the pairwise projection has been segregated in order to determine three different invariant feature matrices as:
\begin{align}
    \mathcal{D}_1 = &  \frac{1}{n^2}\mathcal{G}^{\top} \mathcal{X}\mathcal{X}^{\top} \mathcal{G} \\
    \mathcal{D}_2 = & \frac{1}{n^2}\mathcal{G}^{\top} \mathcal{U}\mathcal{U}^{\top} \mathcal{G} \\
    \mathcal{D}_3 = & \frac{1}{n}\mathcal{G}^{\top} \mathcal{C} 
\end{align}
where $\mathcal{D}_1$ and $\mathcal{D}_2$ are based on the pairwise projection of spatial coordinates and velocity vectors, respectively. Three components $(D_1, D_2, D_3)$ are then concatenated to determine invariant features (replacing the role of $D$ in the original VCNN), which are then mapped to the scalar quantity $\tau$ through the fitting network. Such form closely resembles the determination of rotational invariant edge features (\ref{eq:ri_input}) of GKN as well. 

Since more information is being kept in invariant features with larger dimensions, the number of learnable parameters of the fitting network significantly increases ($\sim$123000) compared to the earlier presented VCNN with $\sim$33000 parameters. Despite the near four-fold increase in model size, the training time per epoch increases only to 4.1$s$ from 3.3$s$. Based on such representation of invariant feature matrix, results have been compared in Table~\ref{fig:vcnn_D_123}. The reduction of prediction error percentage for training and interpolating the testing datasets can be attributed to the increased number of learnable parameters. However, extrapolating testing error percentage is comparable to that of the previously discussed architecture of VCNN.

\begin{table}[tbp]
\centering
\caption{Comparison of predictive performances of two VCNN architectures along with GKN. VCNN ($\mathcal{D}_1$, $\mathcal{D}_2$, $\mathcal{D}_3$) retains more information in input feature matrix $\mathcal{D} = [\mathcal{D}_1 \ \mathcal{D}_2 \ \mathcal{D}_3]^{\top}$ compared to the VCNN ($\mathcal{D}$) architecture (Fig.~\ref{fig:schematic_vcnn}).}
\label{fig:vcnn_D_123}
\begin{tabular}{|| p{3.0cm} | >{\centering\arraybackslash}m{2.5cm} | >{\centering\arraybackslash}m{1.9cm} | >{\centering\arraybackslash}m{1.9cm} | >{\centering\arraybackslash}m{1.9cm} | >{\centering\arraybackslash}m{1.9cm} ||} \hline
 &  & \multicolumn{2}{c|}{
 \begin{tabular}{>{\centering\arraybackslash}m{3.8cm}}
Test Error -- Interpolation \\ \hline
\end{tabular}
 } & \multicolumn{2}{c||}{
\begin{tabular}{>{\centering\arraybackslash}m{3.8cm}}
Test Error -- Extrapolation \\ \hline
\end{tabular}
} \\ 
\quad \ NN Model & Training Error & AOA = 15\degree & AOA = 25\degree & AOA = 5\degree & AOA = 35\degree \\ \ChangeRT{1.1pt}
GKN - with RI & 0.33\% & 1.7\% & 1.4\% & 2.7\% & 2.6\% \\ \hline
VCNN ($D$) & 0.31\% & 3.5\% & 3.2\% & 6.8\% & 5.3\% \\ \hline
VCNN ({\footnotesize $D_1,D_2,D_3$}) & 0.20\% & 2.3\% & 2.1\% & 5.0\% & 5.2\% \\ \hline
\end{tabular}
\end{table}

\section{Neural network architectures details}
\label{architecture}
In this appendix, architecture details of both neural network models are given. VCNN hyperparameters have been adapted as originally proposed by Zhou~et~al.~\cite{zhou2021frame}, for which details for embedding and fitting networks are mentioned separately in Table~\ref{tab:vcnn-details}. To make a fair comparison between both neural network models, we have kept the total number of learnable parameters almost equal for both GKN and VCNN. Since nonlocal mapping from a point cloud or a graph to a scalar quantity $\tau$ is performed by the fitting network for VCNN, only the fitting network of VCNN has been considered for equivalency of total learnable parameters. The Adam optimizer~\cite{kingma2015adam} is adopted to train both of the neural networks. The neural network architectures have been implemented in the machine learning framework PyTorch, while also using PyTorch Geometric for implementing GKN.

\begin{table}[htbp]
\caption{ Architecture details of vector cloud neural network (Fig.~\ref{fig:schematic_vcnn}), which comprises of two neural networks; the embedding network and the fitting network. Neurons in each layer are mentioned in sequence, from the input layer to the hidden layers (if any) and the output layer. The numbers of neurons in the input and output layers are highlighted in bold. Note that the embedding network operates identically on the scalar features $\bm{c}\in \bbR^{l’}$ associated with each point in the cloud and output a row of $m$ elements in matrix~$\cG$.}
\centering
\begin{tabular}{P{5.1cm} P{3.7cm} P{4.5cm}}
\toprule[1pt]
 & Embedding network  & Fitting network ($\mathcal{D} \mapsto \tau$)\\
\midrule
No. of input neurons & $l’ = 7 $ & $m\times m’= 256 \, (\cD \in \bbR^{64\times 4})$\\
No. of hidden layers & 2 & 1 \\
No. of output neurons & $m=64$ & 1 ($\tau \in \bbR$) \\
Neurons in each layer & (\textbf{7}, 32, 64, \textbf{64}) & (\textbf{256}, 128, \textbf{1}) \\
Activation functions & ReLU & ReLU, Linear (last layer) \\
No. of trainable parameters & 19008 & 33025 \\
\bottomrule[1pt]
\end{tabular}
\label{tab:vcnn-details}
\end{table}

\begin{table}[htbp]
\caption{Architecture details of graph kernel network (Fig.~\ref{fig:schematic_vcnn}). Neurons in each layer are mentioned in sequence, from the input layer to the hidden layers (if any) and the output layer. The numbers of neurons in the input and output layers are highlighted in bold. Kernel network operates identically on the invariant edge features $\bm{e}(\bm{q}_i, \bm{q}_j) \in \bbR^d$ associated with each pair of nodes/points in the cloud to output a kernel matrix $\mathcal{K}_{ij}$.}
\centering
\begin{tabular}{P{5.1cm} P{4.5cm}}
\toprule[1pt]
 & Kernel network \\
\midrule
No. of input neurons & $d=16$ \\
No. of hidden layers & $L=2$ \\
No. of output neurons & $m \times m = 256$ ($\mathcal{K}_{ij}\in\bbR^{16\times16}$) \\
Neurons in each layer & (\textbf{16}, 64, 96, \textbf{256}) \\
Activation functions & ReLU, Linear (last layer) \\
No. of trainable parameters & 32577 \\
\bottomrule[1pt]
\end{tabular}
\label{tab:gkn-details}
\end{table}

\clearpage

\bibliographystyle{apalike}

\end{document}